%% file: cvpr_arxiv.tex
\documentclass[final]{cvpr}

\usepackage{times}
\usepackage{epsfig}
\usepackage{graphicx}
\usepackage{amsmath}
\usepackage{amssymb}

\usepackage{booktabs}       
\usepackage{multirow}

\usepackage{graphicx}

\usepackage{xcolor}
\definecolor{CustomGreen}{RGB}{0,140,114}
\definecolor{CustomBurgundy}{RGB}{159,29,53}

\usepackage{amsthm}
\usepackage{bbm}

\usepackage{stfloats}

\usepackage{listings}
\usepackage{upquote}

\usepackage{placeins}

\newcommand\+{\mkern4mu}

\makeatletter

\renewcommand{\paragraph}{%
	\@startsection{paragraph}{4}%
	{\z@}{0.ex plus 0.ex minus .0ex}{-0.5em}{\normalsize\bf}}

\let\originalparagraph\paragraph 
\renewcommand{\paragraph}[2][.]{\originalparagraph{#2#1}}

\makeatother

\usepackage[pagebackref=true,breaklinks=true,colorlinks,bookmarks=false]{hyperref}

\begin{document}
	
	
	\title{Improving Augmentation and Evaluation Schemes for Semantic Image Synthesis
	}
	
	\author{Prateek Katiyar\\
		Bosch Center for Artificial Intelligence\\
		{\tt\small Prateek.Katiyar@de.bosch.com}
		\and
		Anna Khoreva\\
		Bosch Center for Artificial Intelligence\\
		{\tt\small Anna.Khoreva@de.bosch.com}
	}
	
	\maketitle

\input{sections/abstract.tex}
	
	\input{sections/introduction.tex}

	\input{sections/related_work.tex}

	\input{sections/methods.tex}

	\input{sections/evaluation_bias.tex}

	\input{sections/experiments.tex}

	\input{sections/conclusions.tex}
	
	{\small
		\bibliographystyle{ieee_fullname}
		\bibliography{sections/references}
	}
	
	\clearpage

	\setcounter{figure}{0}
	\setcounter{table}{0} 
	
	
	\appendix

\input{cvpr_supplementary_arxiv.tex}

\end{document}

%% file: sections/abstract.tex
\begin{abstract}

Despite data augmentation being a \textit{de facto} technique for boosting the performance of deep neural networks, little attention has been paid to developing augmentation strategies for generative adversarial networks (GANs). To this end, we introduce a novel augmentation scheme designed specifically for GAN-based semantic image synthesis models. We propose to randomly warp object shapes in the semantic label maps used as an input to the generator. The local shape discrepancies between the warped and non-warped label maps and images enable the GAN to learn better the structural and geometric details of the scene and thus to improve the quality of generated images.
While benchmarking the augmented GAN models against their vanilla counterparts, we discover that the quantification metrics reported in the previous semantic image synthesis studies are strongly biased towards specific semantic classes as they are derived via an external pre-trained segmentation network. We therefore propose to improve the established semantic image synthesis evaluation scheme by analyzing separately the performance of generated images on the biased and unbiased classes for the given segmentation network. Finally, we show strong quantitative and qualitative improvements obtained with our augmentation scheme, on both class splits, using state-of-the-art semantic image synthesis models across three different datasets. On average across COCO-Stuff, ADE20K and Cityscapes datasets, the augmented models outperform their vanilla counterparts by $\sim$3 mIoU and $\sim$10 FID points.
\end{abstract}

%% file: sections/introduction.tex
\section{Introduction}
\label{sec:introduction}

\input{figures/teaser.tex}

The reliance of modern deep learning methods on significant amounts of training data has made data augmentation techniques ubiquitous for large-scale machine learning tasks~\cite{Krizhevsky2012ImageNetCW,Szegedy2016RethinkingTI,Ratner2017LearningTC,Taylor2018ImprovingDL,Yun2019CutMixRS}. Besides being used to increase the size and diversity of the training set, data augmentation can also serve as an implicit regularizer and prevent the overfitting of models with high capacity \cite{sajjadi2016regularization,Devries2017ImprovedRO,Berthelot2019MixMatchAH}. 
While in principle augmentation strategies can be applied to any class of models, their utility in improving the training of generative adversarial networks (GANs) for image synthesis has not been widely studied, and very
little attention has been paid to developing novel augmentation strategies specific for GANs.
For the task of semantic image synthesis (SIS), which aims to generate realistic images from the corresponding semantic label maps, recent work has primarily focused on the refinement of conditional GAN architectures \cite{isola2017image,wang2018high,park2019semantic,liu2019learning,tang2019local,Zhu_2020_CVPR}, adopting routine augmentation schemes that were originally designed for image classification~\cite{Krizhevsky2012ImageNetCW}, such as random cropping or flipping.
Although these transformations incorporate geometric invariances about the data domain in the trained models, they are not specifically tuned for the SIS task, and thus do not assist in alleviating its common pitfalls. 

Despite the recent successes of the SIS models~\cite{park2019semantic,liu2019learning,tang2019local,ntavelis2020sesame}, one can still observe unsatisfactory artifacts in the synthesized images, mainly in the generation of object-level fine-grained structures. Since the input label maps do not provide any supervision about the structural content within the semantic segments, the generated images often lack class-relevant structural information and additionally contain undesirable distortions. Fig.~\ref{fig:teaser} shows two such cases, where a recently proposed SIS model~\cite{liu2019learning} adds significant distortions in the building and road segments of the synthesized images, and overall lacks relevant structural details.
Inspired by the task-specific augmentation studies in other vision applications~\cite{dwibedi2017cut,dvornik2018modeling,Singh2017HideandSeekFA,Berthelot2019MixMatchAH,tripathi2019learning}, in this work, we propose an augmentation method specifically designed to overcome the above mentioned limitations of the SIS models.

Our proposed augmentation approach allows to greatly improve the quality of the synthetic images by enabling the generator to focus more on the local shapes and structural details (see Fig.~\ref{fig:teaser}c). We achieve this by randomly warping objects in the semantic label map fed to the SIS model as an input. The local shape discrepancies between the semantic input and the non-warped real image enable the generator to learn geometric properties of the scene better, which may otherwise be ignored since the generator has access to the ground truth scene layout that is perfectly aligned with the real image. Besides the perceptual details, the discriminator also utilizes this misalignment between the real image and the warped label map to distinguish between the real and fake images, forcing the generator to correct distortions introduced to its input by learning the object-level shape details.
To provide an even fine-grained guidance about the class-specific structural details missing from the label map, we add a structure preserving edge loss term to the generator objective. This ensures that the edges in the synthetic image are faithful to those in the corresponding real image.

We demonstrate the efficacy of our augmentation scheme by improving recently proposed SIS models on three different datasets, both quantitatively and qualitatively. Following~\cite{isola2017image,wang2018high,park2019semantic}, besides standard image synthesis metrics, we also use semantic segmentation metrics for the SIS evaluation, which were adopted with a two-fold reasoning. Firstly, a good SIS model should generate images with the layout that closely matches the ground truth semantic label map. Secondly, realistically-looking semantic classes in the generated image should be recognized well by an external semantic segmentation network trained on the real images from an independent dataset. However, we discover that the biases learned by the segmentation network during training leak into the quantification of the synthetic images, resulting in an overestimation of the SIS model's performance (see Fig.~\ref{fig:teaser}, bottom panel). We therefore propose to mitigate this issue by identifying the biased and unbiased classes in all datasets for the given segmentation networks and show the advantages of our proposed augmentation scheme using an extended evaluation on both class splits.

In summary, our main contributions are:
(1) We propose a simple and novel data augmentation strategy designed specifically for GAN-based SIS models. Our augmentation scheme improves the quality of the generated images by encouraging the SIS model to focus more on the local image details and structures.
(2) We showcase a fundamental issue present in the evaluation of SIS models due to the biases learned by the pre-trained segmentation models and propose a fix by extending the SIS model evaluation on dataset-specific biased and unbiased class splits. 
(3) We conduct extensive experiments using state-of-the-art SIS models on three different datasets to demonstrate the efficacy of our proposed augmentation scheme and perform ablation studies to carefully study the effect of augmentation on GAN-based SIS model components.

%% file: figures/teaser.tex
\begin{figure*}[h]
	\begin{center}
		\includegraphics[width=1.0\linewidth]{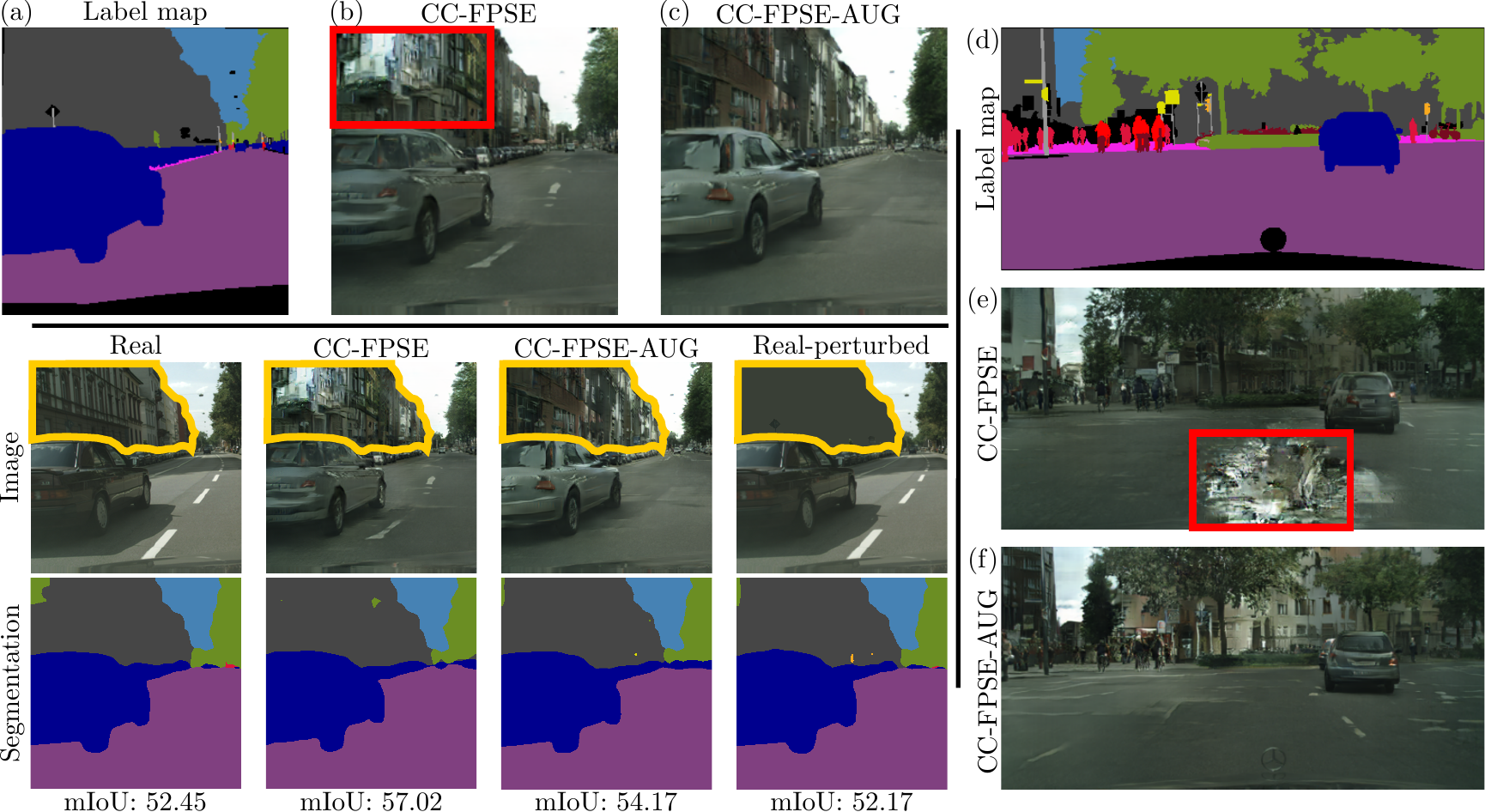}
	\end{center}
	\vspace{-1em}
	\caption{Examples from Cityscapes \cite{cordts2016cityscapes} illustrating the image quality improvements obtained using our proposed augmentation scheme and the adverse effects of segmentation model's bias in the evaluation of SIS models. For clarity, the images in the top and bottom panels have been cropped to size $256 \times 256$. Top panel: label map input (a) and the associated synthetic images generated using the baseline (b) and augmented (c) CC-FPSE \cite{liu2019learning} models. The proposed augmentation scheme fixes the artifact (shown with red box) introduced by the baseline model and improves the overall structural and perceptual details of the synthesized image. Bottom panel: the top row shows real, synthesized (baseline and augmented) and perturbed images for the label map shown in (a), and the bottom row shows the respective image segmentations obtained using a pre-trained segmentation model \cite{drn105} and the mIoU scores. Despite the artifact in the CC-FPSE image, the segmentation model is able to infer the building segment (in yellow), achieving the highest mIoU score. This also applies for the Real-perturbed case, where even after replacing the entire building segment information with its average pixel value, the segmentation model achieves an mIoU score similar to that for the real image. A non-cropped augmentation example is presented in panels (d)-(f).}
	\label{fig:teaser}
	\vspace{-1em}
\end{figure*}

%% file: sections/related_work.tex
\section{Related Work}
\label{sec:related_work}
\paragraph{Semantic image synthesis}

Conditional GANs~\cite{goodfellow2014generative,Mirza2014ConditionalGA} can generate images via side information, such as class labels \cite{Zhang_SAGAN19,Brock2019}, textual descriptions \cite{ReedAYLSL16,Qiao2019MirrorGANLT,Xu2018AttnGANFT,han2017stackgan,ZhangPAMI2018}, images \cite{isola2017image,zhu2017toward,huang2018multimodal}, scene graphs \cite{Johnson2018ImageGF,Tripathi2019UsingSG}, or semantic label maps \cite{park2019semantic,tang2019local,Zhu_2020_CVPR, Zhu_2020_CVPR_SMIS,ntavelis2020sesame} as in this work. 
Using the label maps as a guidance, SIS aims to create photo-realistic images. 
Pix2Pix~\cite{isola2017image} first proposed to use GANs for this task, employing an encoder-decoder generator which takes semantic label map as input and a PatchGAN discriminator. 
Aiming at high-resolution image synthesis, Pix2PixHD~\cite{wang2018high} improved upon \cite{isola2017image} by using multi-scale PatchGAN discriminators and introducing a coarse-to-fine generator architecture. SPADE~\cite{park2019semantic} proposed to fuse the semantic information in the label maps to the generator more effectively via a spatially-adaptive normalization layer. 
To better exploit the semantic maps and adaptively control the synthesis process, CC-FPSE~\cite{liu2019learning} further introduced a
semantics-embedding discriminator and spatially-varying conditional
convolution kernels in the generator.
Most recently, LGGAN~\cite{tang2019local} focused on improving synthesis of small objects and local details, by designing a generator with separate branches that jointly learn the local class-specific and global image-level generation. Likewise, the authors in~\cite{tang2020edge} improve the synthesis of local structures via an attention-based edge guided generator network. 
Besides GANs, CRN~\cite{chen2017photographic} and SIMS~\cite{qi2018semi} train a cascaded refinement convolutional network with a regression loss. 
This work proposes a novel data augmentation scheme for training GAN-based SIS models, which is orthogonal to the previous approaches and can be used for training the above SIS methods. 

\paragraph{Augmentation}

Data augmentation is a widely used technique for generating additional data to train machine learning systems
~\cite{Simard1992EfficientPR,Schlkopf1996IncorporatingII,Krizhevsky2012ImageNetCW,Taylor2018ImprovingDL,Cubuk2019AutoAugmentLA,Tokozume2018BetweenClassLF}. On natural images, the most common form of data augmentation is based on label preserving geometric and photometric transformations \cite{Krizhevsky2012ImageNetCW,Szegedy2016RethinkingTI}, e.g. when images are rotated, scaled, or color-jittered without altering their class labels. 
These transformations are known to suppress the overfitting effects and to improve generalization. 
Recently, label-perturbing data augmentation methods have received a lot of attention~\cite{zhang2018mixup,Verma2018ManifoldMB,Yun2019CutMixRS, harris2020fmix}, proposing to create augmented images by mixing samples from different classes and interpolating their labels accordingly.
Though these approaches have shown to be effective, their potential shortcoming is that the model may learn a biased decision boundary, as the augmented samples are not drawn directly from an underlying distribution. 
Instead of hand-crafted techniques, several works have focused on finding automatic augmentation strategies \cite{Lemley2017SmartAL,Cubuk2019AutoAugmentLA,Cubuk2019RandAugmentPD,Ratner2017LearningTC} or generating samples via GANs~\cite{Zhu2017DataAI,Sixt2018RenderGANGR,Antoniou2017DataAG}. 

Very little attention has been paid to data augmentation policies for GANs themselves, which also benefit from more training data and regularization. For image
augmentations, the mainstream GAN models adopt only random cropping and horizontal flipping as
their augmentation strategy.
A few augmentation techniques have been proposed to improve the training stability of GANs, dealing with the issue of vanishing gradient by adding noise to the images \cite{Sonderby2016AmortisedMI,Arjovsky2017TowardsPM}, blurring images~\cite{SajParMehSch18}, progressively appending the discriminator's input with random bits~\cite{Zhang2018PAGANIG} or
using a wide range of standard augmentations techniques~\cite{Karras2020TrainingGA} to prevent the discriminator from overfitting. Most recently, \cite{Zhao2020ImageAF} provided some guidelines of the effectiveness of various existing augmentation techniques for GAN training. However, the above works mostly focus on the unconditional and class-conditional image synthesis.
In contrast, we propose a novel data augmentation strategy for GAN-based SIS models. Our augmentation scheme improves the quality of the generated samples by encouraging the SIS model to focus more on the local image details and structures.

%% file: sections/methods.tex
\section{Method}
\label{sec:methods}

\input{figures/aug_workflow.tex}

\input{sections/augmentation.tex}

\input{figures/plane_example.tex}
\input{sections/edge_loss.tex}

%% file: figures/aug_workflow.tex
\begin{figure*}[t]
	\vspace{-1em}
	\centering
	\includegraphics[width=0.8\linewidth]{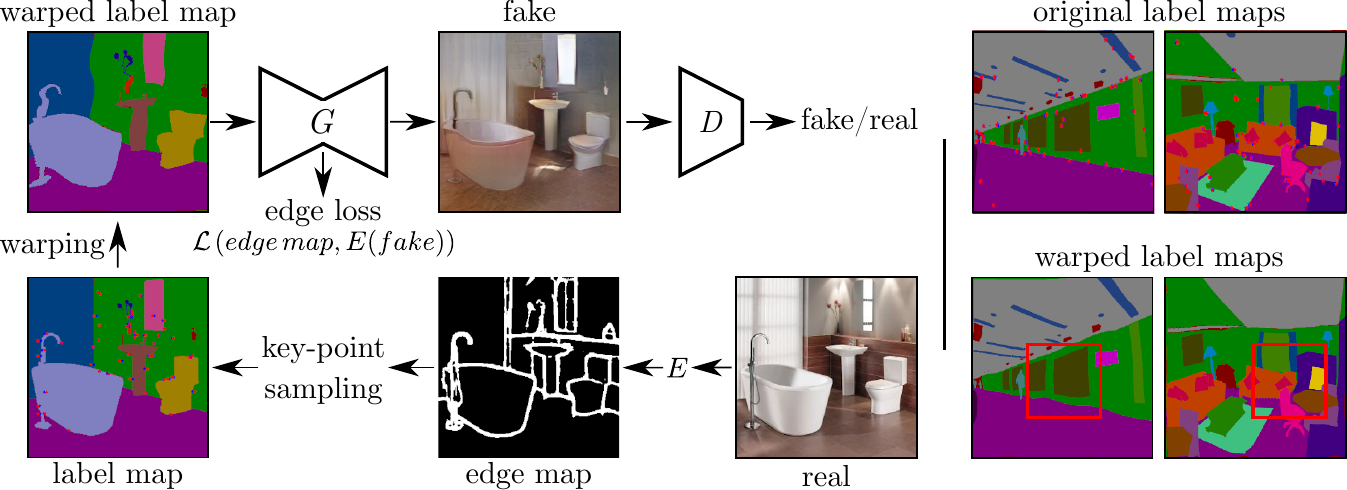}
		
	\caption{Augmented SIS pipeline. Left panel: in contrast to a vanilla model, the input to the generator $G$ in the augmented model is a label map warped based on the edges, estimated by the edge detector $E$. 
		 Not shown is the real image and the warped label map fed into the discriminator $D$. Right panel: examples of original and the respective warped label maps. Zoom in the boxes for details.}
	\vspace{-1em}
	\label{fig:aug_workflow}
\end{figure*}

%% file: sections/augmentation.tex
\subsection{Augmentation}
\label{subsec:augmentation}

In SIS, training dataset consists of a pair of data samples \((s, x)\), where $s$ denotes the input semantic label map\footnote{For certain datasets, SIS models also concatenate instance maps with the label maps.} and $x$ is the respective real image. In this task, the generator is trained to learn the distribution of real images conditioned on the semantic input. Thus, the loss functions for the generator $G$ and the discriminator $D$ take the following form:
\begin{equation}
\begin{aligned}
& \underset{min}{\mathcal{L}_G} = -\mathbb{E}_{s}[\text{log}\+ D(G(s), s)]\text{,}\\
& \underset{max}{\mathcal{L}_D} = \mathbb{E}_{(x, s)}[\text{log}\+ D(x, s)] + \mathbb{E}_{s}[\text{log}\+ (1 - D(G(s), s))]\text{.}
\end{aligned}
\end{equation}
To stabilize the training and improve the fidelity of synthetic images, recent SIS models have also added feature matching and perceptual losses~\cite{wang2018high} to the generator:
\begin{equation}
\begin{aligned}
\underset{min}{\mathcal{L}_G} = & -\mathbb{E}_{s}[\text{log}\+ D(G(s), s)] \\[-5pt]
& + \lambda_{FM} \mathbb{E}_{(x, s)} \mathcal{L}_{FM}(D(x, s), D(G(s), s))\\
& + \lambda_{P} \mathbb{E}_{(x, s)} \mathcal{L}_{P}(F(x), F(G(s)))\text{,}
\end{aligned}
\end{equation}
where $\lambda_{FM}$ and $\lambda_{P}$ are the weights of the respective loss terms, and $F$ is a pre-trained CNN.

While the modifications in the objective functions and model architectures \cite{isola2017image,wang2018high,park2019semantic,liu2019learning,tang2019local} have led to steady improvements in the performance of SIS models, they do not explicitly guide the generator network to learn the local shape details of objects in the real image. In fact, as the generator architectures proposed in the state-of-the-art SIS models~\cite{liu2019learning,park2019semantic} condition the features of all intermediate layers on the semantic input, the generator can simply copy the global structural layout of the scene directly from the conditioning input. This direct dependency may further weaken the generator's ability to learn finer structural details, because the semantic input itself lacks the information about the composition of various classes, e.g. windows/doors for the building class. Thus, to prevent the generator from naively copying the scene layout and encourage it to learn local shape properties of different classes, we propose to warp the objects in the semantic label map fed to the generator. The geometric mismatches between the semantic input and the corresponding real image force the generator to not only learn the perceptual content of different classes, but also to correct the shape distortions introduced to its input. More specifically, we obtain the warped semantic label map $\tilde{s}$ using a transformer function:
$\tilde{s} = t(s)\text{.}$
Here, the thin-plate spline transform $t$ is obtained by estimating the affine and non-affine warping coefficients, which minimize a bending energy function for a set of fixed $\{u, v\}$ and moving $\{\acute{u}, \acute{v}\}$ points~\cite{bookstein1989principal}. To selectively warp the objects in the input semantic label map $s$, we sample the key-points $\{u, v\}$ uniformly from boundary pixels in the edge map (see Fig.~\ref{fig:plane}). Afterwards, the moving points $\{\acute{u}, \acute{v}\}$ are obtained by adding random horizontal and vertical pixel shifts to the previously sampled key-points within a defined range. The amount of pixel shift controls the degree of warping. For each dataset, we determined this parameter experimentally by training multiple models with varying levels of distortions. Fig.~\ref{fig:aug_workflow} shows examples of such warped label maps. We train the augmented models by warping the input label maps from the entire training dataset and conditioning the generator and the discriminator on the warped semantic layouts. The real images fed to the discriminator remain unmodified. During inference, the non-warped semantic label maps are used to generate the synthetic images.

%% file: figures/plane_example.tex
\begin{figure}[h]
	\centering
	\includegraphics[width=1.0\linewidth]{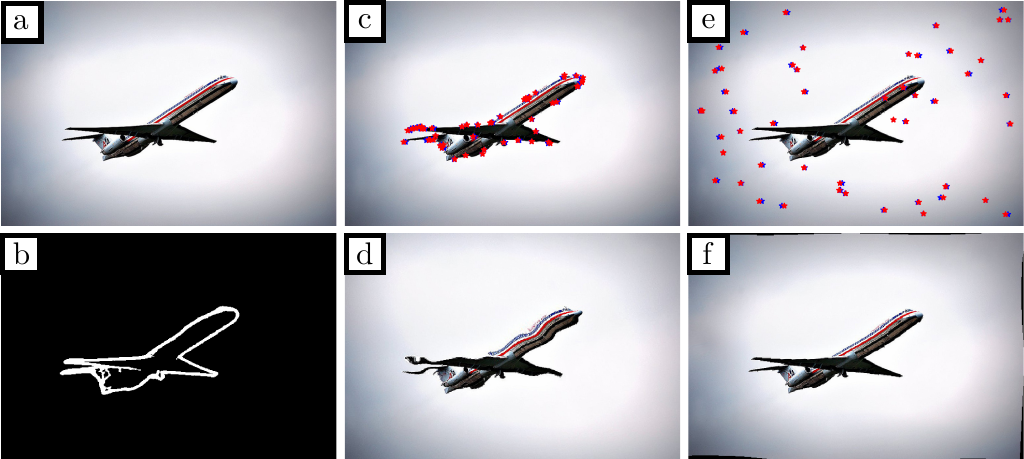}
	\caption{A warping example comparing the effects of sampling key-points from the boundary pixels versus random selection. The first column shows an image of a plane (a) and its edge map extracted using the edge detection network (b). The second column shows key-points sampled from the boundary pixels in the edge map overlaid on the image (c) and the corresponding warped image obtained after applying the thin-plate spline transformer function $t$, estimated using the fixed and moving boundary key-points (d). The last column shows key-points randomly sampled from the image (e) and the respective warped image (f). It is evident from the panels (d) and (f) that compared to random selection, sampling key-points from the boundary pixels in the edge map allows to precisely distort the local shape details of the objects in the image.}
	\label{fig:plane}
	\vspace{-1em}
\end{figure}

%% file: sections/edge_loss.tex
\subsection{Edge Loss}
\label{subsec:edge_loss}

Although warping semantic label maps enables the generator to focus on the local shape details (as described previously), it does not provide class specific fine-grained structural guidance necessary for high fidelity SIS. For example, it is evident from the building class segment (dark gray color) in the label map in Fig. \ref{fig:teaser} that the generator can only infer the class identity and its spatial extent from the semantic input. While the feature matching and perceptual losses aid the generator in learning class specific content, they do not provide explicit structural clues about the fine-grained scene details. 
We therefore propose to add a structure preserving edge loss \cite{khoreva2021} to the generator's output using the edge maps extracted by the edge detector $E$ for the label map warping. 
We implement the edge loss $\mathcal{L}_{E}$ as the L2 difference between the edge maps extracted from the synthetic and the real images: $\mathcal{L}_{E} = \lVert E(x) - E(G(\tilde{s})) \rVert_2$.
Jointly, the augmentation and the edge loss complement each other, as they allow the generator to focus on different aspects of the target scene geometry. The final $G$ and $D$ losses of the augmented SIS model with the edge loss are: 
\begin{equation}
\begin{aligned}
\underset{min}{\mathcal{L}_G} = & -\mathbb{E}_{s}[\text{log}\+ D(G(\tilde{s}), \tilde{s})] \\[-5pt]
& + \lambda_{FM} \mathbb{E}_{(x, s)} \mathcal{L}_{FM}(D(x, \tilde{s}), D(G(\tilde{s}), \tilde{s}))\\
& + \lambda_{P} \mathbb{E}_{(x, s)} \mathcal{L}_{P}(F(x), F(G(\tilde{s}))) \\[-2pt]
& + \lambda_{E} \mathbb{E}_{(x, s)} \mathcal{L}_{E}(E(x), E(G(\tilde{s})))\text{,}
\end{aligned}
\end{equation}\vspace{-10pt}
\begin{align*}
\underset{max}{\mathcal{L}_D} = \mathbb{E}_{(x, s)}[\text{log}\+ D(x, \tilde{s})] + \mathbb{E}_{s}[\text{log}\+ (1 - D(G(\tilde{s}), \tilde{s}))]\text{.}
\end{align*}
An illustrative overview of the complete augmentation model is presented in Fig.~\ref{fig:aug_workflow}.

%% file: sections/evaluation_bias.tex
\section{Evaluation Bias}
\label{sec:evaluation_bias}

The use of semantic segmentation metrics for evaluating the quality of synthetic images has become a common practice in SIS \cite{isola2017image,wang2018high,park2019semantic,liu2019learning}. Such evaluation is based on the hypothesis that realistically looking synthetic images can be classified correctly by a semantic segmentation network that was originally trained on real images~\cite{isola2017image}. Thus, we first introduce the preliminaries associated with SIS evaluation. Thereafter, we present the image perturbations with which we determine the semantic classes that lead to a biased evaluation of SIS models for a given segmentation network.

Let \(X \in \mathbb{R}^{H \times W}\) and \(Y \in \{1, 2..., N_{cl}\}^{H \times W}\) denote an input image and its densely labelled semantic map with $N_{cl}$ categories. Here, $H$ and $W$ represent the image height and width. For simplicity we restrict the discussion below to single channel images, although it is fully generalizable to multi-channel images. Given an arbitrary image input $X$ to a pre-trained segmentation network, let $Y_{pred}$ be the predicted segmentation map. Using $Y$ and $Y_{pred}$, we can calculate the following evaluation metrics for each class $i$: Pixel Accuracy \((PA_i) = n_{ii}/t_{i}\) and, Intersection over Union $(IoU_i) = n_{ii}/(t_{i} \text{+} \sum_{j} n_{ji} \text{--} n_{ii})$. Here, $n_{ji}$ and $t_i$ are the number of pixels of class $j$ that are labelled as class $i$ in $Y_{pred}$ and the total number of pixels of class $i$ in $Y$~\cite{long2015fully}.

\input{figures/perturbation.tex}

To assess whether the segmentation model is biased towards specific semantic classes, we modify $X$ by using different perturbation schemes and evaluate the metrics defined above by feeding the perturbed images into the segmentation network. As contextual information~\cite{hoyer2019grid,mottaghi2014role} and statistics~\cite{wang2016learnable} have been shown to strongly influence the predictions of a semantic segmentation model, we aim to utilize the segmentations of perturbed images to determine classes that can be correctly inferred by the segmentation network even when the information within the class segment is artificially altered. 

For \( (u, v) \in (\{1, 2,..., H\}, \{1, 2,..., W\}) \), let \(M_i(u, v) = \mathbbm{1}\{Y(u, v) = i\} \) be the mask for class $i$, where $\mathbbm{1}$ is the indicator function. We can define the perturbed image $\tilde{X}_i$ for class $i$ as: \(\tilde{X}_i(M_i, X) = X \circ (1-M_i) + P \circ M_i\). Here, \(P \in \mathbb{R}^{H \times W}\) is the applied perturbation and $\circ$ denotes the Hadamard product. Since SIS relies on pre-trained segmentation networks to evaluate the fake image pool, our choice of perturbations include approaches that replace the class segment information with the statistics derived from the segment itself. The following perturbations (constant, average, Gaussian blur and lognormal) are a basic approximation of a SIS model's ability to perform label-to-image translation:
\begin{equation}
	P(u, v) = \begin{cases}
		c_{0}\\[-5pt]
		\dfrac{1}{\sum_{u}\sum_{v}M(u, v)}\sum_{u}\sum_{v} M(u, v) * X(u, v)\\[8pt]
		G(\sigma_0) \circledast X \\
		{\raise.17ex\hbox{$\scriptstyle\mathtt{\sim}$}}\+ lognormal(\mu, \sigma)
	\end{cases}\label{eq:perturbation_schemes}
\end{equation}

where $c_0$ is a fixed grayscale value, $\sigma_0$ is the standard deviation parameter of the Gaussian kernel, and the mean $\mu$ and standard deviation $\sigma$ are determined from the masked image segment. In our experiments, we determine $\sigma_0$ for each dataset by taking into account the class-wise segment area statistics across the entire validation set. The constant perturbation additionally covers classes that can be identified by a segmentation model solely based on the class silhouette and the neighboring context. Examples of all four perturbation schemes are presented in Fig.~\ref{fig:perturbation}.   

By feeding the perturbed images into the segmentation model, we obtain the set $\mathbb{M}_{p_i}$ that contains the aforementioned metrics calculated for class $i$ for all perturbation schemes. Given the score \(m_i \in \{PA_i, IoU_i\}\) for the original image, the class $i$ is considered as biased if the following criterion is met for any of the two respective perturbed metric sets:
\begin{equation}
	i = 
	\begin{cases}
		\text{\textbf{biased}} & \text{if } \exists\+ m_{p_i} \ni m_{p_i} > \delta * m_i, \forall\+ m_{p_i} \in \mathbb{M}_{p_i} \text{,} \\
		\text{\textbf{unbiased}} & \text{otherwise.}
	\end{cases}	\label{eq:bias_criterion}
\end{equation}
Where the factor $\delta * m_i$ determines the threshold for the perturbed metrics for class $i$ to be regarded as biased. It is considered as biased because the segmentation network is able to identify the class segment in $\tilde{X}_i$ with sufficiently high accuracy for any of the four perturbation schemes. In practice, we group classes in the biased and unbiased splits using the cumulative $PA_i$ and $IoU_i$ metrics calculated for the entire validation set. We chose $\delta=2/3$ and additionally compute the following metrics to evaluate SIS models across both splits: \(\text{Mean Accuracy ($MA_{s}$)} = 1/N_{cl_{s}} \sum_{i} PA_i\) and \(\text{Mean IoU ($mIoU_{s}$)} = 1/N_{cl_{s}} \sum_{i} IoU_i\), where $N_{cl_{s}}$ denotes the number of classes in the split $s$.

%% file: figures/perturbation.tex
\begin{figure*}[t]
	\begin{center}
	\includegraphics[width=1.0\linewidth]{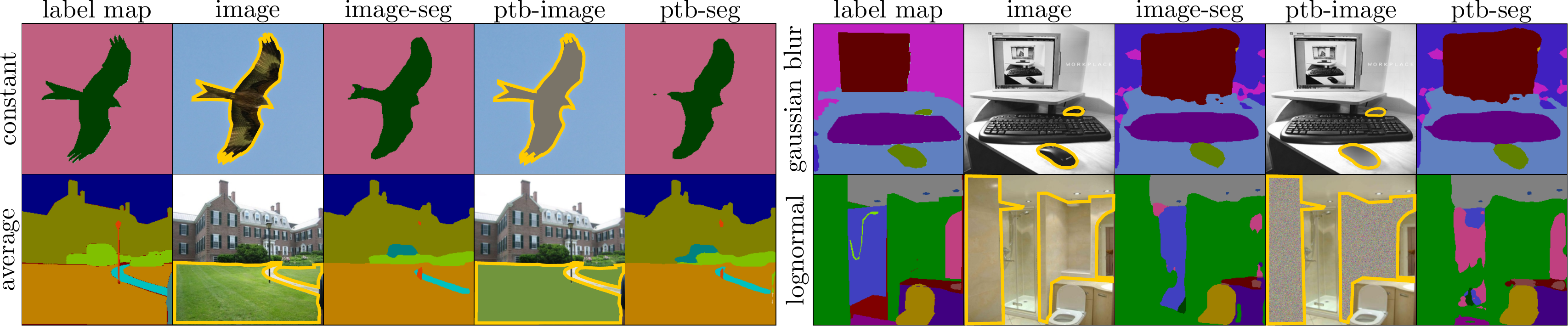}
	\end{center}
	\vspace{-0.5em}
	\caption{Image perturbation examples exhibiting the bias present in pre-trained segmentation models. The perturbed class segments are highlighted with yellow contours. For each perturbed image (ptb-image), the segmentation model is able to correctly infer the semantic category of the affected segment (as in image-seg), despite the corruption of perceptual and structural details. More details in Sec.~\ref{subsec:results_evaluation_bias}.}
	\label{fig:perturbation}
	\vspace{-1em}
\end{figure*} 

%% file: sections/experiments.tex
\section{Experiments}
\label{sec:experiments}

\input{sections/datasets_models_and_implementation_details.tex}

\input{figures/qualitative.tex}

\input{sections/results_evaluation_bias.tex}

\input{sections/results_augmentation.tex}

%% file: sections/datasets_models_and_implementation_details.tex
\subsection{Datasets, Models and Implementation Details}
\label{subsec:datasets_and_models}

\paragraph{Datasets}
We perform experiments on COCO-Stuff \cite{caesar2018coco}, ADE20K \cite{zhou2017scene} and Cityscapes \cite{cordts2016cityscapes}. Each dataset provides pixel-level annotations for all images. COCO-Stuff includes natural scene images with 172 classes, divided into 118k training and 5k validation samples. Likewise, ADE20K contains 20k training and 2k validation images with a mixture of indoor and outdoor scenes with 150 categories. Cityscapes comprises 5k training and 500 validation driving scene images with 35 classes.

\paragraph{Baselines}
As baselines, we choose three recently proposed GAN-based SIS models: Pix2PixHD \cite{wang2018high}, SPADE \cite{park2019semantic} and CC-FPSE \cite{liu2019learning}. 
We compare the qualitative and quantitative performances of the baselines with the respective augmented models, where we warp the semantic input and add the edge loss. We denote these models with -AUG suffix.

\paragraph{Segmentation and edge detection models}
For consistent quantitative evaluations with the published baselines, we use the following segmentation models: DeepLabV2 (COCO-Stuff) \cite{chen2017deeplab,deeplabv2}, UperNet101 (ADE20K) \cite{xiao2018unified,upernet101}, and  DRN-D-105 (Cityscapes) \cite{yu2017dilated,drn105}. In addition to mIoU, PA and MA metrics defined in Sec. \ref{sec:evaluation_bias}, we calculate F\'rechet Inception Distance (FID) \cite{heusel2017gans} to compare the distributions of real and fake image samples. We use a pre-trained Bi-Directional Cascade Network~\cite{he2019bi,bdcn} to estimate edges.

\paragraph{Implementation details}
For a fair comparison, we use the original code provided by the authors \cite{ccfpse,spade,pix2pixhd}. We train each model using 4 NVIDIA Tesla V100 GPUs. Except the batch sizes (provided in the supp. data), all other settings were kept to the defaults of the respective models. The resolution of the synthesized images are $256\times256$ for COCO-Stuff and ADE20K, and $512\times256$ for Cityscapes. We use $\lambda_{E}$ of 10 for the edge loss. The Cityscapes and ADE20K models are trained for 200 epochs, whereas the COCO-Stuff models are trained for 100 epochs. We retrain each baseline model from scratch for all datasets.

%% file: figures/qualitative.tex
\begin{figure*}[t]
	\centering
	\begin{minipage}[c]{0.9\linewidth}
		\centering
		\setlength{\tabcolsep}{0em}
		\renewcommand{\arraystretch}{0}
		
		\hfill{}%
		\begin{tabular}{@{}c@{}c@{}c@{}c@{}c@{}c@{}}
			
			\centering
			
			Label map & Real image & SPADE & SPADE-AUG & CC-FPSE & CC-FPSE-AUG \vspace{0.05cm} \tabularnewline
			
			\includegraphics[width=0.166\linewidth, height=0.09\textheight]{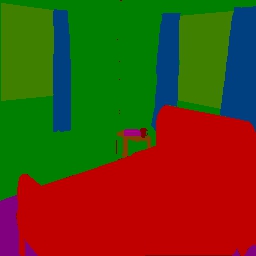} & {\footnotesize{}}
			\includegraphics[width=0.166\linewidth, height=0.09\textheight]{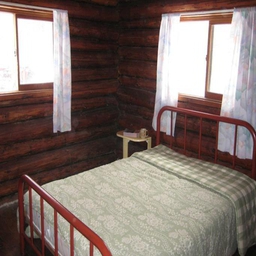} & {\footnotesize{}}
			\includegraphics[width=0.166\linewidth, height=0.09\textheight]{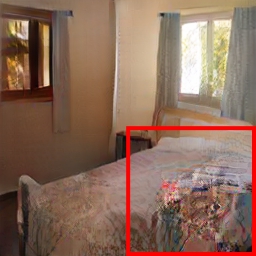} & {\footnotesize{}}
			\includegraphics[width=0.166\linewidth, height=0.09\textheight]{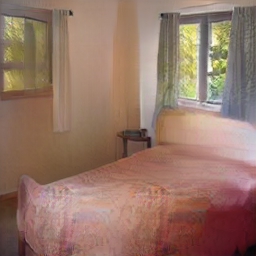} & {\footnotesize{}}
			\includegraphics[width=0.166\linewidth, height=0.09\textheight]{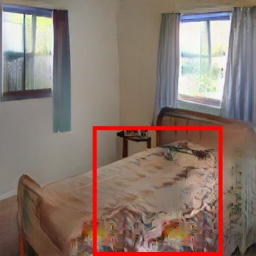} & {\footnotesize{}}
			\includegraphics[width=0.166\linewidth, height=0.09\textheight]{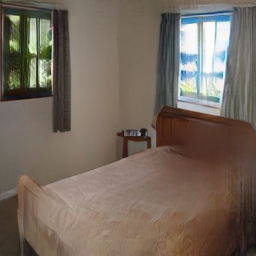} \vspace{0.05cm} \tabularnewline
			
			\includegraphics[width=0.166\linewidth, height=0.09\textheight]{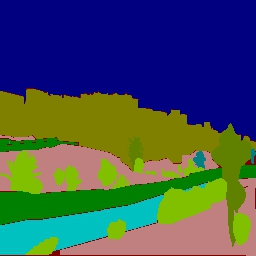} & {\footnotesize{}}
			\includegraphics[width=0.166\linewidth, height=0.09\textheight]{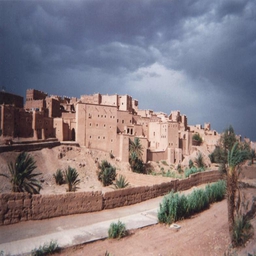} & {\footnotesize{}}
			\includegraphics[width=0.166\linewidth, height=0.09\textheight]{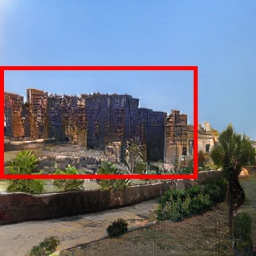} & {\footnotesize{}}
			\includegraphics[width=0.166\linewidth, height=0.09\textheight]{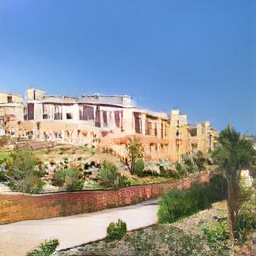} & {\footnotesize{}}
			\includegraphics[width=0.166\linewidth, height=0.09\textheight]{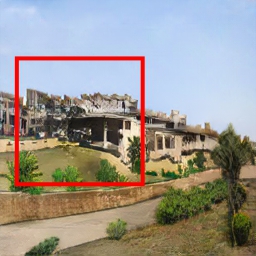} & {\footnotesize{}}
			\includegraphics[width=0.166\linewidth, height=0.09\textheight]{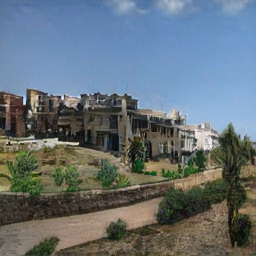} \vspace{0.05cm} \tabularnewline
			
			\includegraphics[width=0.166\linewidth, height=0.09\textheight]{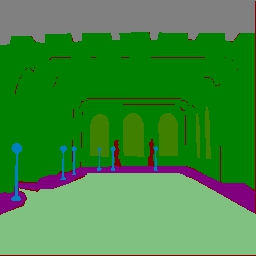} & {\footnotesize{}}
			\includegraphics[width=0.166\linewidth, height=0.09\textheight]{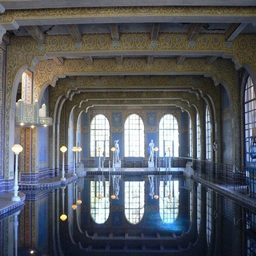} & {\footnotesize{}}
			\includegraphics[width=0.166\linewidth, height=0.09\textheight]{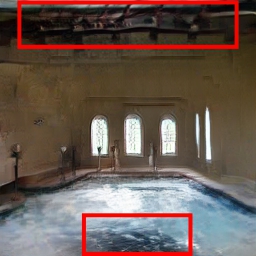} & {\footnotesize{}}
			\includegraphics[width=0.166\linewidth, height=0.09\textheight]{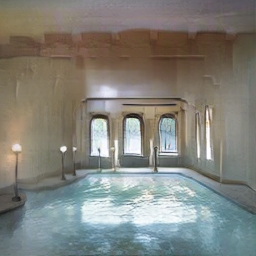} & {\footnotesize{}}
			\includegraphics[width=0.166\linewidth, height=0.09\textheight]{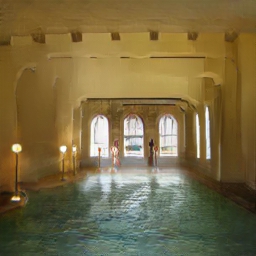} & {\footnotesize{}}
			\includegraphics[width=0.166\linewidth, height=0.09\textheight]{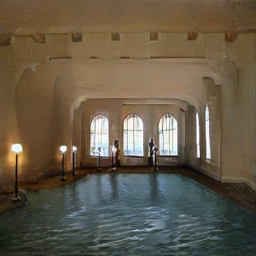} \vspace{0.05cm} \tabularnewline
			
		\end{tabular}
	
	\end{minipage}
	\vspace{0.2em}
	\caption{ADE20K results using SPADE and CC-FPSE baselines, and the augmented models. The red boxes highlight image artifacts.} 
	\label{fig:ade20k_qualitative}
	\vspace{-0.25em}
\end{figure*}

%% file: sections/results_evaluation_bias.tex
\subsection{Evaluation Bias Results}
\label{subsec:results_evaluation_bias}

\paragraph{Biased Classes}

Following Sec.~\ref{sec:evaluation_bias}, we first identify the biased and unbiased classes in all datasets. To this end, we apply the pre-trained models on the perturbed validation set images created for each class in the respective dataset. We also separately test the models on the corresponding real validation set images. Based on the perturbed and unperturbed cumulative validation set metrics ($PA_i$ and $IoU_i$), we group the classes in the biased and unbiased categories. Note that we focus only on strongly biased classes. For a $\delta$ value of $2/3$ in Eq.~\ref{eq:bias_criterion}, we find 29, 52 and 5 biased classes in COCO-Stuff, ADE20K and Cityscapes (provided in the supp. data). Smaller values of $\delta$ dilute the biased class split. 

Fig.~\ref{fig:perturbation} shows examples of original and perturbed image segmentations for all four perturbation schemes defined in Eq.~\ref{eq:perturbation_schemes}. For the constant perturbation image, the segmentation network is able to identify the bird class almost entirely without leveraging any class-specific texture. Likewise, the average perturbation example shows that the average color and (quite likely) the positioning of the grass segment are sufficient for the segmentation model to achieve a near-perfect classification. In the Gaussian blur perturbation example, the network correctly identifies the mouse in front of the keyboard, while misses the one behind. This example shows the effect of context bias picked up by the segmentation model, as in majority of training images the mouse is placed either to the front or alongside the keyboard. The final lognormal perturbation example shows bidirectional effects of the bias learned by the segmentation network during training. Here, the perturbed wall segment is classified accurately, whereas the unaltered shower-door segment is misclassified as mirror. These examples highlight that for the biased classes, even unrealistic SIS model images may lead to high evaluation metrics. We investigate this in the next section by performing evaluations on the synthetic images generated using all three baselines.

\paragraph{Analysis on Baselines}

In Table~\ref{tab:bias} we report the results of evaluating the synthetic images of all baselines and the real images across all datasets. To our surprise--when considering all classes--the segmentation networks perform better on the synthetic images than on the real images for COCO-Stuff and ADE-20K. The reason for better segmentation performance on the synthetic images becomes clear, as we bifurcate the metrics into the biased and unbiased classes. While overall the segmentation networks perform better on the biased classes than the unbiased ones, this performance gap is significantly higher for the synthetic images generated using the SIS baselines compared to the real images, especially for COCO-Stuff and ADE-20K. We also notice large differences between Pix2PixHD and the other baselines, indicating that newer architectures improve upon the synthesis of the unbiased classes. Having verified the adverse effects of the segmentation network's bias in the evaluation of SIS models, in the next section we extend the analysis of the augmentation results to both class splits.

\input{tables/bias.tex}

%% file: tables/bias.tex
\begin{table*}[t]
	
	\setlength{\tabcolsep}{0.25em}
	\renewcommand{\arraystretch}{1.0}
	\centering
	
		\begin{tabular}{@{}l@{\hspace{0.4cm}}cccc@{\hspace{0.3cm}}cccc@{\hspace{0.3cm}}cccc@{}}
			
			\multicolumn{1}{c}{\multirow{2}{*}{}} & \multicolumn{4}{c}{ mIoU (all classes) $\uparrow$} &  \multicolumn{4}{c}{ $\Delta$  mIoU$_{BC}$} & \multicolumn{4}{c}{ $\Delta$  mIoU$_{UC}$} \tabularnewline
			\cmidrule(l){2-5} \cmidrule(l){6-9} \cmidrule(l){10-13} 
			\multicolumn{1}{c}{} &  \small Real &  \small CC-FPSE &  \small SPADE &  \small Pix2PixHD &  \small Real &  \small CC-FPSE &  \small SPADE &  \small Pix2PixHD &  \small Real &  \small CC-FPSE &  \small SPADE &  \small Pix2PixHD \tabularnewline
			
				COCO-Stuff       &  35.4 &  \textbf{41.0} &  37.8 &  12.0 &  \textcolor{CustomGreen}{+3.9} &  \textcolor{CustomGreen}{+6.3} &  \textcolor{CustomGreen}{+5.7} &  \textcolor{CustomGreen}{+9.3} &  \textcolor{CustomBurgundy}{-0.8} &  \textcolor{CustomBurgundy}{-1.3} &  \textcolor{CustomBurgundy}{-1.1} &  \textcolor{CustomBurgundy}{-1.9} \tabularnewline
			
					ADE20K     &  36.5 &  \textbf{42.6} &  39.6 &  24.4 &  \textcolor{CustomBurgundy}{-0.2} &  \textcolor{CustomGreen}{+1.9} &  \textcolor{CustomGreen}{+2.1} &  \textcolor{CustomGreen}{+3.5} &  \textcolor{CustomGreen}{+0.1} &  \textcolor{CustomBurgundy}{-1.0} &  \textcolor{CustomBurgundy}{-1.0} &  \textcolor{CustomBurgundy}{-1.8} \tabularnewline
			
					Cityscapes &  \textbf{62.0} &  61.8 &  59.2 &  56.7 &  \textcolor{CustomGreen}{+17.4} &  \textcolor{CustomGreen}{+17.5} &  \textcolor{CustomGreen}{+20.7} &  \textcolor{CustomGreen}{+22.1} &  \textcolor{CustomBurgundy}{-6.2} &  \textcolor{CustomBurgundy}{-6.3} &  \textcolor{CustomBurgundy}{-7.3} &  \textcolor{CustomBurgundy}{-7.9} \tabularnewline[0.1cm] 
			
			\multicolumn{1}{c}{\multirow{2}{*}{}} &  \multicolumn{4}{c}{ MA (all classes) $\uparrow$} &  \multicolumn{4}{c}{  $\Delta$  MA$_{BC}$} &  \multicolumn{4}{c}{ $\Delta$  MA$_{UC}$} \tabularnewline\cmidrule(l){2-5} \cmidrule(l){6-9} \cmidrule(l){10-13} 
			
					COCO-Stuff       &  47.2 &  \textbf{52.4} &  48.8 &  18.0 &  \textcolor{CustomGreen}{+5.2} &  \textcolor{CustomGreen}{+9.1} &  \textcolor{CustomGreen}{+8.1} &  \textcolor{CustomGreen}{+10.7} &  \textcolor{CustomBurgundy}{-1.1} &  \textcolor{CustomBurgundy}{-1.9} &  \textcolor{CustomBurgundy}{-1.7} &  \textcolor{CustomBurgundy}{-2.2} \tabularnewline
			
					ADE20K     &  44.7 &  \textbf{49.8} &  46.6 &  29.5 &  \textcolor{CustomBurgundy}{-0.3} &  \textcolor{CustomGreen}{+2.9} &  \textcolor{CustomGreen}{+2.5} &  \textcolor{CustomGreen}{+5.2} &  \textcolor{CustomGreen}{+0.1} &  \textcolor{CustomBurgundy}{-1.5} &  \textcolor{CustomBurgundy}{-1.3} &  \textcolor{CustomBurgundy}{-2.7} \tabularnewline
			
					Cityscapes &  \textbf{72.3} &  71.6 &  68.3 &  65.4 &  \textcolor{CustomGreen}{+13.4} &  \textcolor{CustomGreen}{+14.0} &  \textcolor{CustomGreen}{+17.2} &  \textcolor{CustomGreen}{+18.8} &  \textcolor{CustomBurgundy}{-4.8} &  \textcolor{CustomBurgundy}{-5.0} &  \textcolor{CustomBurgundy}{-6.2} &  \textcolor{CustomBurgundy}{-6.7} \tabularnewline
		\end{tabular}
	\vspace{0.5em}
	\caption{Metrics highlighing the bias leaked from the segmentation network in the quantification of the SIS models. For clarity, the $\Delta$ values for the biased classes ($BC$) and unbiased classes ($UC$) in \textcolor{CustomGreen}{green}/\textcolor{CustomBurgundy}{red} depict an increment/decrement from the corresponding scores obtained with all classes.} \label{tab:bias}
	\vspace{-1.25em}
\end{table*}

%% file: sections/results_augmentation.tex
\subsection{Augmentation Results}
\label{subsec:results_augmentation}

\paragraph{Quantitative results}

We report quantitative results of the baselines and the respective augmented models in Table~\ref{tab:aug}. Here we show only FID and mIoU metrics. The full table including PA and MA metrics is provided in the supp. data. 
Even without optimizing the weight for the edge loss, we see that the augmented models outperform the corresponding baselines on each dataset. As the augmented generators focus more on the local shape and structural details, we particularly see gains on the unbiased class metrics. In some cases, the overall gain is only contributed by the improvements on the unbiased classes, e.g. CC-FPSE mIoU on COCO-Stuff and SPADE mIoU on Cityscapes. In fact, for each model pair and dataset combination in Table~\ref{tab:aug}, the augmented models show consistent gains over the baselines on the overall and unbiased mIoU (this also applies to MA metric) indicating that the general fine-grained improvements of the augmentation extend across a variety of GAN-based SIS architectures as well as datasets containing diverse scenes and objects. Adding only the edge loss to the baselines improves the model performance, but not as much as the full augmentation scheme (see the supp. data). 
The CC-FPSE and SPADE baselines trained on COCO-Stuff and ADE20K achieve lower FID scores than the respective augmented models. However, for both cases the augmented models perform better on almost all other metrics. We notice that the proposed class splits provide an additional level of granularity when benchmarking SIS models, as they allow to identify cases of pseudo-improvements where the overall boost in performance is caused mainly due to a gain in the biased class metric. For instance on Cityscapes, PA of SPADE (93.1) is higher than that of CC-FPSE (92.8)\footnote{Following DRN-D-105~\cite{drn105}, unlike \cite{park2019semantic,liu2019learning}, we ignore the unlabelled class in the Cityscapes evaluation.}. But, this improvement is only caused by the biased classes. On the unbiased classes, CC-FPSE (84.5) outperforms SPADE (82.3) by more than 2 points.

\input{tables/aug.tex}

\paragraph{Qualitative results}

The qualitative examples comparing SPADE and CC-FPSE baselines, and the augmented models are shown in Fig.~\ref{fig:ade20k_qualitative}. Here, we illustrate that the proposed augmentation scheme reduces distortions, adds fine-grained structural details and enhances the perceptual realism of the synthetic images. The first 2 rows exemplify the distortions commonly present in the synthetic images of both baselines by the red boxes (Cityscapes examples shown in Fig.~\ref{fig:teaser}). We found that the proposed augmentation significantly reduce such irregularities. Moreover, the images in the second row show how the augmented models gradually add local structural details to the synthetic images and produce a high fidelity image. The final row shows that compared to the baselines, the augmented models also improve the overall perceptual realism of the translated images. Additional examples (including the results from the edge loss baselines) supporting these findings are provided in the supp. data.

\paragraph{Ablations}

The results of ablations on the CC-FPSE model on Cityscapes are provided in Table~\ref{tab:ablations}. For other SIS models and datasets, the reader is referred to the supp. data. We first compare our proposed augmentation technique with the CutMix augmentation~\cite{yun2019cutmix}, which has recently been used to regularize unconditional GAN training~\cite{schonfeld2020u}. While the CutMix variant improves upon the baseline model on mIoU metrics, it remains inferior to the proposed (task-specific) augmentation approach. Next, to understand which component of the SIS model benefits most with the proposed augmentation, we selectively augment only the generator (Gonly) or discriminator (Donly) input label maps. We observe that while augmenting both networks results in the best performance, augmenting the generator's input is most critical for a GAN-based SIS model. Finally, to confirm the utility of both the label map warping and the edge loss in our proposed approach, we train one additional augmented model on each dataset in which we remove the edge loss (-AUG-woEL). This ablation confirms complementary benefits of the full scheme (-AUG), as adding the edge loss on top of the warping leads on average to a better performance (decreasing FID and increasing mIoU$_{UC}$).     


\input{tables/ablations.tex}

%% file: tables/aug.tex
\begin{table*}[t]
	\setlength{\tabcolsep}{0.05em}
	\renewcommand{\arraystretch}{1.0}
	
	\centering
		\begin{tabular}{@{}l@{\hspace{0.2cm}}cccc@{\hspace{0.5cm}}cccc@{\hspace{0.5cm}}cccc@{}}
			
		  \multirow{2}{*}{ Model }& \multicolumn{4}{c}{ COCO-Stuff} & \multicolumn{4}{c}{ ADE20K} & \multicolumn{4}{c}{ Cityscapes} \tabularnewline  
			
		&	 FID $\downarrow$&  mIoU $\uparrow$&  mIoU$_{BC}$&  mIoU$_{UC}$ &  FID $\downarrow$&  mIoU $\uparrow$&  mIoU$_{BC}$&  mIoU$_{UC}$ & FID $\downarrow$&  mIoU $\uparrow$&  mIoU$_{BC}$&  mIoU$_{UC}$
			\tabularnewline  
			
			\cmidrule(lr){2-5}  	\cmidrule(lr){6-9}  	\cmidrule(lr){10-13} 
			
		  CC-FPSE &  \textbf{18.9} &  41.0 &  \textbf{47.3} &  39.7  &  33.2 &  42.6 &  44.5 &  41.6 &  53.6 &  61.8&  79.3&  55.5  \tabularnewline

		  CC-FPSE-AUG &  19.1 & 	\textbf{42.1}&  46.3&  \textbf{41.2} &  \textbf{32.6} &  \textbf{44.0} &  \textbf{45.8} &  \textbf{43.1} &  \textbf{52.1}&  \textbf{63.1}&  \textbf{80.1}&  \textbf{57.0}\tabularnewline[0.1cm]  
			
		  SPADE &  \textbf{22.5}&  37.8 &  43.5 &  36.7 &  \textbf{34.4} &  39.6 &  41.7 &  38.6 &  64.7&  59.2&  \textbf{79.9}&  51.9 \tabularnewline
		   SPADE-AUG&  22.7&  \textbf{38.2}&  43.5 &  \textbf{37.1} &  34.6&  \textbf{41.2} &  \textbf{43.2} &  \textbf{40.2} &  \textbf{62.3}&  \textbf{60.4}&  79.8&  \textbf{53.5}\tabularnewline[0.1cm] 
			
		 Pix2PixHD &  128.7 &  12.0 &  21.2 &  10.1 &  59.1 &  24.4 &  27.9 &  22.6 &  73.6&  56.7&  78.8&  48.8 \tabularnewline 
		 Pix2PixHD-AUG &  \textbf{54.2}&  \textbf{21.9}&  \textbf{31.8}&  \textbf{19.9} &  \textbf{41.5} &  \textbf{32.5} &  \textbf{36.0} &  \textbf{30.7} &  \textbf{72.7}&  \textbf{58.0}&  \textbf{79.1}&  \textbf{50.5}  \tabularnewline
		 
		 	\cmidrule(lr){2-5}  	\cmidrule(lr){6-9}  	\cmidrule(lr){10-13} 
		 	
		 Average $\Delta$ & 24.7  &  3.8 & 3.2  &  3.9 & 6  & 3.7  & 3.6  &  3.7 &  1.6&  1.3&  0.3 &  1.6 \tabularnewline

		\end{tabular}%
	\vspace{0.5em}
	\caption{Evaluation comparison across datasets. ${BC}$ and ${UC}$ denote the biased and unbiased class splits. \textbf{Bold} indicates the best model between the baseline and its augmented variant (-AUG).}
	\label{tab:aug}
	\vspace{-1.25em}
\end{table*}

%% file: tables/ablations.tex
\begin{table}[t]
	
	\setlength{\tabcolsep}{0.07em}
	\renewcommand{\arraystretch}{1.0}
	
	\centering
		\begin{tabular}{@{}l@{\hspace{0.2cm}}cccc@{}}
			
		  \multirow{2}{*}{ Model } & \multicolumn{4}{c}{ Cityscapes} \tabularnewline  
			
		&	 FID $\downarrow$&  mIoU $\uparrow$&  mIoU$_{BC}$&  mIoU$_{UC}$
			\tabularnewline  
			
		  \cmidrule(lr){2-5}  	
			
		  CC-FPSE & 53.6 & 61.8 & 79.3 & 55.5 \tabularnewline [0.1cm]
		  
		  CC-FPSE-CutMix & 55.2 & 62.0 & 78.6 & 56.1 \tabularnewline [0.1cm]  	 
		  
		  CC-FPSE-Donly & 77.7 & 53.6 & 74.7 & 46.0 \tabularnewline
		  
		  CC-FPSE-Gonly & 57.6 & 55.5 & 75.9 & 48.2 \tabularnewline [0.1cm]  
		  
		  CC-FPSE-AUG-woEL & 53.4 & \textbf{63.1} & \textbf{80.4} & 56.9 \tabularnewline [0.1cm]  
		  
		  CC-FPSE-AUG &  \textbf{52.1}&  \textbf{63.1} & 80.1 & \textbf{57.0} \tabularnewline
		 
		\end{tabular}%
	\vspace{0.5em}
	\caption{Ablations on Cityscapes. ${BC}$ and ${UC}$ denote the biased and unbiased class splits. \textbf{Bold} indicates the best model. For Donly and Gonly models, the augmentation was applied only to the discriminator and the generator input, respectively.  Augmented model without the edge loss is denoted as AUG-woEL.}
	\label{tab:ablations}
	\vspace{-1.5em}
\end{table}

%% file: sections/conclusions.tex
\section{Conclusions}
\label{sec:conclusions}

In this work, we propose a novel data augmentation method for GAN-based SIS models. Targeting the shortcomings of the recent SIS studies, the proposed augmentation scheme allows to significantly improve the overall quality of the synthetic images, in particular the local shapes and structures inside the semantic classes. In addition, to better analyze and benchmark the improvements of the SIS models, we extend the semantic segmentation metrics used for the SIS evaluation. Specifically, to mitigate the adverse effects of biases picked up by the segmentation network, we split the evaluation of semantic classes into biased and unbiased groups. Enabled by this new analysis, we observe that the conventional SIS models strongly underperform on unbiased classes, while our proposed augmentation method improves their results on both class groups.

%% file: cvpr_supplementary_arxiv.tex
\section*{Supplementary Material}
This supplementary material is structured as follows:
\begin{description}
	\item \ref{sec:biased_classes}: Additional information on biased semantic classes across datasets.
	\item \ref{sec:exp_details}: A summary of the network architectures and the training details.
	\begin{description}
		\item Supplementary Table~\ref{tab:batch_sizes}: Batch sizes used in all experiments.
	\end{description}
	\item \ref{sec:quant_results}: Additional quantitative results.
	\begin{description}
		\item Supplementary Table~\ref{tab:aug_full}: Full evaluation comparison of the baselines with the edge loss and the augmented variants across datasets.
		\item Supplementary Table~\ref{tab:cutmix_selective_augmentation_ablations}: Complete ablation results including CutMix, Donly and Gonly model performances across datasets.
		\item Supplementary Table~\ref{tab:edge_loss_ablation}: Performance of the augmented models without the edge loss.
		\item Supplementary Table~\ref{tab:matching_aware_loss_ablation}: Performance of the augmented models with the matching aware GAN loss.
	\end{description}
	\item \ref{sec:qual_results}: Additional qualitative results.
	\begin{description}
		\item Supplementary Figures~\ref{fig:ccfpse_coco}--\ref{fig:pix2pixhd_cityscapes}: Visual comparison of the images generated using our method with those of the baselines and the edge loss models. 
	\end{description}
\end{description}

\input{sections/biased_classes.tex}
\input{sections/supp_experimental_details.tex}
\input{sections/supp_quantitative_results.tex}
\input{sections/supp_qualitative_results.tex}

%% file: sections/biased_classes.tex
\section{Biased Classes}
\label{sec:biased_classes}

In this section, we detail the list of classes that are identified as biased using the perturbation approach presented in Sec. \ref*{sec:evaluation_bias}. For a $\delta$ value of 2/3, we identify 29, 52 and 5 biased classes in COCO-Stuff, ADE20K and Cityscapes, respectively. The Gaussian perturbation parameter  $\sigma_0$ for COCO-Stuff, ADE20K and Cityscapes was set to 25.0, 35.0 and 27.0, respectively. The parameter $\sigma_0$ is a function of the kernel size $K$ ($\sigma_0 = 1/3 * K$), which was detemined by calculating the average class area statistics for all semantic categories in the dataset. The biased class names and their associated ids are provided below in a dictionary format.

\subsection*{COCO-Stuff}
\begin{lstlisting}
13:  "stop sign",
16:  "bird",
28:  "umbrella",
34:  "frisbee",
35:  "skis",
36:  "snowboard",
41:  "skateboard",
42:  "surfboard",
49:  "knife",
62:  "chair",
74:  "mouse",
86:  "vase",
101: "carpet",
110: "desk-stuff",
115: "floor-other",
120: "fog",
124: "grass",
126: "ground-other",
135: "mountain",
142: "plant-other",
144: "platform",
145: "playingfield",
149: "road",
154: "sand",
157: "sky-other",
159: "snow",
167: "textile-other",
172: "wall-concrete",
173: "wall-other"
\end{lstlisting}

\subsection*{ADE20K}
\begin{lstlisting}
1:   "wall",
3:   "sky",
4:   "floor flooring",
5:   "tree",
6:   "ceiling",
7:   "road route",
8:   "bed ",
10:  "grass",
12:  "sidewalk pavement",
14:  "earth ground",
16:  "table",
17:  "mountain mount",
18:  "plant flora plant life",
20:  "chair",
29:  "rug carpet carpeting",
31:  "armchair",
34:  "desk",
37:  "lamp",
40:  "cushion",
41:  "base pedestal stand",
42:  "box",
43:  "column pillar",
47:  "sand",
51:  "refrigerator icebox",
53:  "path",
65:  "coffee table cocktail table",
69:  "hill",
71:  "countertop",
74:  "kitchen island",
76:  "swivel chair",
83:  "light light source",
85:  "tower",
86:  "chandelier pendant pendent",
87:  "awning sunshade sunblind",
94:  "pole",
96:  "bannister banister balustrade 
      balusters handrail",
107: "canopy",
115: "tent collapsible shelter",
126: "pot flowerpot",
129: "lake",
131: "screen silver screen 
      projection screen",
132: "blanket cover",
134: "hood exhaust hood",
136: "vase",
138: "tray",
139: "ashcan trash can garbage can 
      wastebin ash bin ash-bin 
      ashbin dustbin trash barrel 
      trash bin",
140: "fan",
141: "pier wharf wharfage dock",
142: "crt screen",
143: "plate",
144: "monitor monitoring device",
148: "glass drinking glass"
\end{lstlisting}

\subsection*{Cityscapes}
\begin{lstlisting}
7:   "road",
11:  "building",
17:  "pole",
21:  "vegetation",
23:  "sky"
\end{lstlisting}

%% file: sections/supp_experimental_details.tex
\section{Experimental Details}
\label{sec:exp_details}

\subsection*{Network Architectures}

As we use the implementations provided by the authors for all experiments conducted in this study, we refer the reader to the original source code and articles for complete details of the generator and discriminator network architectures~\cite{ccfpse,spade,pix2pixhd,wang2018high,park2019semantic,liu2019learning}. In our experiments, we found that Pix2PixHD adds significant distortions to the generated images in the form of vertical and horizontal stripes. We also observe significant (unsatisfactory) variations in the performance of Pix2PixHD model and its variants. We tried improving Pix2PixHD baseline by adding spectral normalization~\cite{miyato2018spectral} to both generator and discriminator networks. However, this modification only helped in removing the artifacts from the Cityscapes baseline. Therefore, for all Cityscapes experiments in this study we use the spectral norm variant of the original Pix2PixHD model. 

\subsection*{Training Details}

In addition to the baseline and augmented models, we also train strong baselines by adding the edge loss to the vanilla baseline models. We denote the strong baselines with -EL suffix. The learning objective of Pix2PixHD for the adversarial loss is least squares as proposed in~\cite{mao2017least}. Whereas, SPADE and CC-FPSE use the Hinge loss formulation~\cite{lim2017geometric}. Pix2PixHD and SPADE use a weight of 10 for both feature matching and perceptual losses. CC-FPSE uses a weight of 20 for the feature matching loss and a weight of 10 for the perceptual loss. We do not train SPADE and CC-FPSE models with the image encoder option. All models use ADAM~\cite{adamopt} optimizer. The batch sizes used in all experiments are provided in Supplementary Table~\ref{tab:batch_sizes}. SPADE and CC-FPSE use a learning rate of $0.0001$ and $0.0004$ for the generator and discriminator, respectively. Pix2PixHD uses a common learning rate of $0.0002$ for both networks. All models linearly decay the learning rate to zero from epoch 100 to 200 for Cityscapes and ADE20K experiments. To determine the appropriate warping regime, we train multiple models with varying levels of maximum pixel shifts. For each model, we sample the displacement values between the fixed and moving points from a uniform distribution $\mathcal{U}$ as following: \({\raise.17ex\hbox{$\scriptstyle\mathtt{\sim}$}} \+ \mathcal{U}(-a,a)\). Where $a$ denotes the maximum pixel shift. We found that a maximum pixel shift of 4 units works well for almost all the models. The average statistic of $a$ across the best augmented models is $4.11\pm1.05$.

\input{tables/batch_sizes.tex} 

%% file: tables/batch_sizes.tex
\begin{table}[h]
	
	\setlength{\tabcolsep}{0.07em}
	\renewcommand{\arraystretch}{1.0}
	
	\centering

		\begin{tabular}{@{}l@{\hspace{0.2cm}}c@{\hspace{0.3cm}}c@{\hspace{0.3cm}}c@{}}

		  \multirow{2}{*}{ Model } & \multicolumn{3}{c}{ Dataset} \tabularnewline  
			
		  &	COCO-Stuff & ADE20K &  Cityscapes
			\tabularnewline  
		  
			
		  \cmidrule(lr){2-4}  	
			
		  CC-FPSE & 16 & 16 & 8 \tabularnewline [0.1cm]
		  
		  SPADE & 48 & 48 & 32 \tabularnewline [0.1cm]
		  
		  Pix2PixHD & 48 & 48 & 16 \tabularnewline 
		 
		\end{tabular}%
	\vspace{0.5em}
	\caption{Batch sizes used for all experiments detailed in the main text and the supplementary data. Depending on the model and the dataset, the batch size was selected as the largest input that could fit in the GPU memory.}
	\label{tab:batch_sizes}
\end{table}

%% file: sections/supp_quantitative_results.tex
\section{Quantitative Results}
\label{sec:quant_results}

In Supplementary Table~\ref{tab:aug_full}, we provide the full quantitative evaluations of the baselines, the strong baselines (baselines with the edge loss) and the respective augmented models. Overall, we notice that the edge loss and the augmented variants provide incremental improvements over the baseline models. On the unbiased classes, with the exception of SPADE-EL model's PA$_{UC}$ metric on COCO-Stuff (66.3), the augmented models perform better than their edge loss counterparts across all datasets and metrics. In fact, for this exception too, MA$_{UC}$ metric of the augmented model (47.7) is higher than that of the edge loss model (47.1), suggesting more general benefits of the full augmentation scheme across all COCO-Stuff classes. The CC-FPSE-EL model achieves lower FID score than the respective augmented model on COCO-Stuff, however, the latter model performs better on almost all other metrics. We observe that the newer architectures benefit more with the addition of the edge loss. For example, CC-FPSE-EL model overall performs better than the respective baseline across all datasets, however, for SPADE we only see clear improvements on ADE20K and mixed results on Cityscapes. Whereas Pix2PixHD-EL model shows strong gains over the baseline model on COCO-Stuff, the opposite holds true for ADE20K. On Cityscapes, the results are mixed similar to SPADE. The augmented models, however, show consistent increments in evaluation metrics across all models and datasets. 

The full results of CutMix and selective augmentation (-Donly and -Gonly) ablations are provided in Supplementary Table~\ref{tab:cutmix_selective_augmentation_ablations}. The CutMix augmentation~\cite{yun2019cutmix} has recently been shown to be effective for image classification. The authors in~\cite{schonfeld2020u} further employ CutMix to regularize GAN training. We augment the baseline SIS models using CutMix by following the approach presented in~\cite{schonfeld2020u}. More specifically, to prevent the discriminator from learning a biased decision boundary, we set the class label for the CutMix image as fake. The results in Supplementary Table~\ref{tab:cutmix_selective_augmentation_ablations} clearly show that for each model, our task-specific augmentation approach (-AUG) outperforms CutMix across all datasets. While in some cases, the -CutMix model variants achieve lower FID scores than the -AUG counterparts (CC-FPSE-CutMix on COCO-Stuff and SPADE-CutMix on ADE20K), they perform poorly on all other metrics.

The purpose of selective augmentation was to individually study the effects of the proposed augmentation approach on the generator and discriminator networks. Therefore, in this experiment, we selectively augment only the generator (Gonly) or discriminator (Donly) input label maps. For Donly ablation, we additionally remove the edge loss term from the generator objective to account for increments/decrements in the performance arising from augmenting the discriminator network only. In the case of Donly augmentation, both fake and real images are misaligned from the input label map fed to the discriminator due to the applied warping. Whereas, in the case of Gonly augmentation, the fake image fed to the discriminator is misaligned from the real image and the respective label map input, as long as the generator is unable to fully correct the distortions introduced to its input. From the results presented in Supplementary Table~\ref{tab:cutmix_selective_augmentation_ablations}, we observe that overall augmenting both networks leads to the best performance. However, we notice that on an average augmenting the generator is more effective than augmenting the discriminator. These findings imply that in the case of Gonly augmentation, the misalignment between the fake and real images as well the label map input fed to the discriminator allow the discriminator to better distinguish between the real and fake samples, thus forcing the generator to learn fine grained structural details in order to correct the distortions introduced to its input.         

Supplementary Table~\ref{tab:edge_loss_ablation} shows the results for the augmented models trained without the edge loss. We train one such ablation model for each dataset, whichever is fastest to train depending on the dataset size. The main objective of this ablation experiment is to assess the contributions of both--the edge loss and warping--components in the full augmentation scheme. While the CC-FPSE augmented model without the edge loss yields semantic segmentation scores comparable to those of the corresponding augmented model, its FID increases by 1.3 points. The SPADE and Pix2PixHD augmented models without the edge loss perform worse than their original augmented variants on all metrics (except for SPADE-AUG-woEL FID), suggesting the complementary benefits of both components in the full augmentation scheme. Since the edge loss provides additional structural guidance to the generator for the details that are missing from the semantic segments in the input label maps, its combination with the proposed warping scheme provides complementary advantages to the augmented SIS models.

Supplementary Table~\ref{tab:matching_aware_loss_ablation} presents the evaluation for the augmented models with the matching aware loss. The intuition behind adding the matching aware loss is to utilize the unwarped semantics by letting the discriminator of the augmented SIS model classify the real image and warped semantics pair as fake and the respective pair with the unwarped semantics as real. If the discriminator is able to successfully distinguish between these two pairs, then it should be able to provide a highly fine-grained geometric guidance to the generator. Such an auxiliary pair has been shown to improve the performance of conditional GANs previously~\cite{ReedAYLSL16}. We train the augmented variants of all three baselines with the matching aware loss on ADE20K dataset. For all models, we do not see any improvements in adding the matching aware loss on top of the proposed augmentation scheme.

%% file: sections/supp_qualitative_results.tex
\section{Qualitative Results}
\label{sec:qual_results}

In Supplementary Figures~\ref{fig:ccfpse_coco}--\ref{fig:pix2pixhd_cityscapes}, we show several image examples comparing the baseline and its edge loss and augmented variants for all three SIS models in a dataset-wise manner. In each figure, we also highlight image areas with distortions using red boxes. Consistent with the quantitative evaluation, the augmented models greatly improve the fidelity of the generated images. We notice that the augmentation substantially minimizes the distortions commonly present in the images synthesized using the vanilla baseline models. Furthermore, as the edge loss and warping provide additional structural guidance to the generator, we observe clear fine-grained improvements within the semantic classes in the generated images of the augmented models compared to those of the baselines. Whereas the edge loss models improve over the baselines on both the aforementioned factors to some extent, these improvements are not consistent across all models and datasets, as is the case for the augmented models. Lastly, we also notice that the images generated by the augmented models are perceptually more consistent than the samples generated using the respective baselines and the edge loss models. We hypothesize that these perceptual improvements are a result of the better overlap between the supports of the real and fake image distributions. Our interpretation is supported by previous evidence~\cite{Sonderby2016AmortisedMI}, where the authors increase the overlap between the supports of the two distributions by means of simple augmentation approaches, such as adding Gaussian noise to both real and fake samples. 

\clearpage

\input{tables/aug_full.tex}

\input{tables/ablation_cutmix_selective_aug.tex}

\input{tables/ablation_edge_loss.tex}

\input{tables/matching_aware_loss.tex}

\input{figures/ccfpse_coco.tex}

\input{figures/ccfpse_ade20k.tex}

\input{figures/ccfpse_cityscapes.tex}

\input{figures/spade_coco.tex}

\input{figures/spade_ade20k.tex}

\input{figures/spade_cityscapes.tex}

\input{figures/pix2pixhd_coco.tex}

\input{figures/pix2pixhd_ade20k.tex}

\input{figures/pix2pixhd_cityscapes.tex}

%% file: tables/aug_full.tex
\begin{table*}[h]
	\setlength{\tabcolsep}{0.2em}
	\renewcommand{\arraystretch}{0.95}

	\centering
	\resizebox{0.9\linewidth}{!}{%
		\begin{tabular}{@{}lc@{\hspace{0.3cm}}c@{\hspace{0.7cm}}ccc@{\hspace{0.7cm}}ccc@{\hspace{0.7cm}}ccc@{}}
			
			& \small  Method& \small FID $\downarrow$& \small mIoU $\uparrow$& \small mIoU$_{BC}$& \small mIoU$_{UC}$& \small PA $\uparrow$& \small PA$_{BC}$& \small PA$_{UC}$& \small MA $\uparrow$& \small MA$_{BC}$& \small MA$_{UC}$ \tabularnewline  
			\cmidrule(lr){2-12} 
			
			\multirow{9}{*}{  \rotatebox{90}{COCO-Stuff }} & \small \small CC-FPSE & \small 18.9 & \small 41.0 & \small \textbf{47.3} & \small 39.7 & \small 71.0 & \small 77.3 & \small 68.4 & \small 52.4 & \small 61.5 & \small 50.5 \tabularnewline
			& \small CC-FPSE-EL & \small \textbf{18.6} & \small 41.8 & \small 46.4 & \small 40.9 & \small 71.1 & \small 76.7 & \small 68.7 & \small 52.8 & \small 61.2 & \small 51.1 \tabularnewline
			& \small CC-FPSE-AUG & \small 19.1 & \small	\textbf{42.1} & \small 46.3 & \small \textbf{41.2} & \small \textbf{71.5} & \small \textbf{78.1} & \small \textbf{68.8} & \small \textbf{53.1} & \small \textbf{61.8} & \small \textbf{51.3} \tabularnewline[0.2cm]  
			
			& \small \small SPADE   & \small \textbf{22.5} & \small 37.8 & \small \textbf{43.5} & \small 36.7 & \small 68.6 & \small 74.5 & \small 66.2 & \small 48.8 & \small 56.9 & \small 47.1 \tabularnewline
			& \small SPADE-EL & \small 22.8 & \small 37.8 & \small 43.4 & \small 36.7 & \small	68.3 & \small 73.1 & \small \textbf{66.3} & \small 48.9 & \small \textbf{57.6} & \small 47.1 \tabularnewline 
			& \small \small SPADE-AUG & \small 22.7 & \small \textbf{38.2} & \small \textbf{43.5} & \small \textbf{37.1} & \small \textbf{68.8} & \small \textbf{75.3} & \small 66.1 & \small \textbf{49.4} & \small 57.4 & \small \textbf{47.7} \tabularnewline[0.2cm] 
			
			& \small Pix2PixHD & \small 128.7 & \small 12.0 & \small 21.2 & \small 10.1 & \small 36.9 & \small 47.6 & \small 32.5 & \small 18.0 & \small 28.7 & \small 15.8 \tabularnewline
			& \small Pix2PixHD-EL & \small 91.2 & \small 17.9 & \small 27.3 & \small 15.9 & \small 43.9 & \small 55.7 & \small 39.0 & \small 25.5 & \small 37.7 & \small 23.0 \tabularnewline
			& \small 	Pix2PixHD-AUG& \small \textbf{54.2} & \small \textbf{21.9} & \small \textbf{31.8} & \small \textbf{19.9} & \small \textbf{54.1} & \small \textbf{69.2} & \small \textbf{47.8} & \small \textbf{29.5} & \small \textbf{46.8} & \small \textbf{26.0} \tabularnewline
			
			
			\cmidrule(lr){2-12}	
			
			\multirow{9}{*}{  \rotatebox{90}{ADE20K }} & \small \small	CC-FPSE   & \small 33.2 & \small 42.6 & \small 44.5 & \small 41.6 & \small 82.2 & \small 87.6 & \small 72.0 & \small 49.8 & \small 52.7 & \small 48.3 \tabularnewline
			& \small CC-FPSE-EL & \small 34.2 & \small 43.1 & \small 44.7 & \small 42.2 & \small 81.7 & \small 87.5 & \small 70.8 & \small 50.4 & \small 53.4 & \small 48.8 \tabularnewline	
			& \small CC-FPSE-AUG & \small \textbf{32.6} & \small \textbf{44.0} & \small \textbf{45.8} & \small \textbf{43.1} & \small \textbf{83.0} & \small \textbf{88.2} & \small \textbf{73.5} & \small \textbf{51.3} & \small \textbf{54.5} & \small \textbf{49.6} \tabularnewline[0.2cm] 
			
			& \small SPADE & \small \textbf{34.4} & \small 39.6 & \small 41.7 & \small 38.6 & \small 81.0 & \small 85.8 & \small 72.1 & \small 46.6 & \small 49.1 & \small 45.3 \tabularnewline
			& \small SPADE-EL & \small 34.9 & \small 40.8 & \small 42.2 & \small 40.1 & \small \textbf{81.3} & \small \textbf{86.2} & \small 72.2 & \small 47.7 & \small 50.0 & \small \textbf{46.5} \tabularnewline
			& \small SPADE-AUG & \small 34.6 & \small \textbf{41.2} & \small \textbf{43.2} & \small \textbf{40.2} & \small \textbf{81.3} & \small 86.0 & \small \textbf{72.5} & \small \textbf{48.0} & \small \textbf{50.8} & \small \textbf{46.5} \tabularnewline[0.2cm] 
			
			& \small Pix2PixHD & \small 59.1 & \small 24.4 & \small 27.9 & \small 22.6 & \small 71.8 & \small 80.5 & \small 55.3 & \small 29.5 & \small 34.7 & \small 26.8 \tabularnewline
			& \small Pix2PixHD-EL & \small 65.2 & \small 20.4 & \small 24.1 & \small 18.4 & \small 69.1 & \small 77.8 & \small 52.9 & \small 25.0 & \small 30.8 & \small 22.0 \tabularnewline
			& \small Pix2PixHD-AUG & \small \textbf{41.5} & \small \textbf{32.5} & \small \textbf{36.0} & \small \textbf{30.7} & \small \textbf{77.9} & \small \textbf{84.4} & \small \textbf{65.9} & \small \textbf{38.2} & \small \textbf{42.7} & \small \textbf{35.8} \tabularnewline
			
			
			\cmidrule(lr){2-12}  		
			
			\multirow{9}{*}{ \rotatebox{90}{Cityscapes }} & \small CC-FPSE & \small 53.6 & \small 61.8 & \small 79.3 & \small 55.5 & \small 92.8 & \small 94.6 & \small 84.5 & \small 71.6 & \small 85.6 & \small 66.6 \tabularnewline
			& \small CC-FPSE-EL & \small 54.7 & \small	63.0 & \small \textbf{80.1} & \small 56.9 & \small 93.4 & \small \textbf{95.4} & \small 84.5 & \small 72.0 & \small \textbf{85.9} & \small 67.0 \tabularnewline		
			& \small CC-FPSE-AUG & \small \textbf{52.1} & \small \textbf{63.1} & \small \textbf{80.1} & \small \textbf{57.0} & \small \textbf{93.5} & \small \textbf{95.4} & \small \textbf{85.2} & \small \textbf{72.4} & \small 85.6 & \small \textbf{67.7} \tabularnewline[0.2cm]  
			
			& \small SPADE & \small 64.7 & \small 59.2 & \small \textbf{79.9} & \small 51.9 & \small 93.1 & \small \textbf{95.6} & \small 82.3 & \small 68.3 & \small \textbf{85.5} & \small 62.1 \tabularnewline
			& \small SPADE-EL & \small 62.8 & \small 59.4 & \small 79.8 & \small 52.1 & \small \textbf{93.2} & \small 95.5 & \small 82.8 & \small 68.2 & \small 85.4 & \small 62.0 \tabularnewline
			& \small SPADE-AUG & \small \textbf{62.3} & \small \textbf{60.4} & \small 79.8 & \small \textbf{53.5} & \small \textbf{93.2} & \small 95.5 & \small \textbf{83.0} & \small \textbf{69.7} & \small 85.2 & \small \textbf{64.2} \tabularnewline[0.2cm] 
			
			& \small Pix2PixHD & \small 73.6& \small 56.7& \small 78.8& \small 48.8 & \small \textbf{92.7} & \small \textbf{95.3} & \small 80.7 & \small 65.4 & \small 84.2 & \small 58.7 \tabularnewline
			& \small Pix2PixHD-EL & \small 74.8 & \small 56.6 & \small 78.6 & \small 48.8 & \small 92.4 & \small 95.2 & \small 80.4 & \small 65.9 & \small 84.2 & \small 59.4 \tabularnewline	
			& \small Pix2PixHD-AUG & \small \textbf{72.7} & \small \textbf{58.0} & \small \textbf{79.1} & \small \textbf{50.5} & \small \textbf{92.7} & \small \textbf{95.3} & \small \textbf{81.2} & \small \textbf{66.9} & \small \textbf{84.6} & \small \textbf{60.5} \tabularnewline 
		\end{tabular}%
		}
		\vspace{0.5em}
		\caption{Evaluation comparison across datasets. ${BC}$ and ${UC}$ denote the biased and unbiased class splits. \textbf{Bold} indicates the best model among the baseline and its edge loss (-EL) and augmented (-AUG) variants. For the ease of comparison, we repeat the metrics from the baselines and the augmented models from the main text. The augmented models refer to the baseline models trained with the full augmentation scheme proposed in this work (warping + edge loss).}
		\label{tab:aug_full}
\end{table*}

%% file: tables/ablation_cutmix_selective_aug.tex
\begin{table*}[h]
	\setlength{\tabcolsep}{0.2em}
	\renewcommand{\arraystretch}{0.95}
	
	\centering
	\resizebox{0.9\linewidth}{!}{%
		\begin{tabular}{@{}lc@{\hspace{0.3cm}}c@{\hspace{0.7cm}}ccc@{\hspace{0.7cm}}ccc@{\hspace{0.7cm}}ccc@{}}
			
			& \small  Method& \small FID $\downarrow$& \small mIoU $\uparrow$& \small mIoU$_{BC}$& \small mIoU$_{UC}$& \small PA $\uparrow$& \small PA$_{BC}$& \small PA$_{UC}$& \small MA $\uparrow$& \small MA$_{BC}$& \small MA$_{UC}$ \tabularnewline  
			\cmidrule(lr){2-12} 
			
			
			\multirow{15}{*}{  \rotatebox{90}{COCO-Stuff }} & \small \small CC-FPSE & \small \textbf{18.9} & \small 41.0 & \small \textbf{47.3} & \small 39.7 & \small 71.0 & \small 77.3 & \small 68.4 & \small 52.4 & \small 61.5 & \small 50.5 \tabularnewline
			
			& \small CC-FPSE-CutMix & \small 19.0 & \small 41.1 & \small 45.6 & \small 40.2 & \small 70.4 & \small 76.7 & \small 67.8 & \small 52.1 & \small 60.5 & \small 50.4 \tabularnewline
			
			& \small CC-FPSE-Donly & \small 21.1 & \small 40.2 & \small	43.8 & \small 39.5 & \small	70.0 & \small 75.5 & \small	67.7 & \small 51.3 & \small 58.8 & \small	49.7 \tabularnewline
			& \small CC-FPSE-Gonly & \small 20.7 & \small 40.6 & \small	43.8 & \small 39.9 & \small	70.3 & \small 75.1 & \small	68.3 & \small 52.0 & \small 59.3 & \small 50.5 \tabularnewline
			
			& \small CC-FPSE-AUG & \small 19.1 & \small	\textbf{42.1} & \small 46.3 & \small \textbf{41.2} & \small \textbf{71.5} & \small \textbf{78.1} & \small \textbf{68.8} & \small \textbf{53.1} & \small \textbf{61.8} & \small \textbf{51.3} \tabularnewline[0.2cm]  
			
		    
			& \small \small SPADE   & \small \textbf{22.5} & \small 37.8 & \small \textbf{43.5} & \small 36.7 & \small 68.6 & \small 74.5 & \small 66.2 & \small 48.8 & \small 56.9 & \small 47.1 \tabularnewline
			
			& \small SPADE-CutMix & \small 22.8 & \small	38.0 & \small	43.6 & \small	36.9 & \small 68.5 & \small 73.9 & \small	\textbf{66.3} & \small 49.0 & \small \textbf{57.4} & \small	47.2 \tabularnewline 
			
			& \small SPADE-Donly & \small 23.2 & \small	38.0 & \small	42.8 & \small	37.0 & \small	68.7 & \small \textbf{75.3} & \small	65.9 & \small	48.8 & \small	57.2 & \small	47.1 \tabularnewline 
			& \small SPADE-Gonly & \small 23.5 & \small	\textbf{38.2} & \small 43.3 & \small \textbf{37.1} & \small	68.5 & \small	74.9 & \small	65.9 & \small	49.1 & \small 57.2 & \small	47.5 \tabularnewline 
			
			& \small \small SPADE-AUG & \small 22.7 & \small \textbf{38.2} & \small \textbf{43.5} & \small \textbf{37.1} & \small \textbf{68.8} & \small \textbf{75.3} & \small 66.1 & \small \textbf{49.4} & \small \textbf{57.4} & \small \textbf{47.7} \tabularnewline[0.2cm] 
			
			
			& \small Pix2PixHD & \small 128.7 & \small 12.0 & \small 21.2 & \small 10.1 & \small 36.9 & \small 47.6 & \small 32.5 & \small 18.0 & \small 28.7 & \small 15.8 \tabularnewline
			
			& \small Pix2PixHD-CutMix & \small 123.4 & \small	14.7 & \small	24.5 & \small	12.7 & \small	40.9 & \small	56.7 & \small	34.3 & \small	21.8 & \small	35.3 & \small	19.0 \tabularnewline
			
			& \small Pix2PixHD-Donly & \small 55.2 & \small	21.3 & \small	30.4 & \small	19.5 & \small	53.7 & \small	67.7 & \small	47.9 & \small 29.1 & \small 45.4 & \small 25.7 \tabularnewline
			& \small Pix2PixHD-Gonly & \small \textbf{50.0} & \small \textbf{22.5} & \small \textbf{31.9} & \small \textbf{20.6} & \small \textbf{55.1} & \small \textbf{69.9} & \small \textbf{49.0} & \small \textbf{30.5} & \small 45.8 & \small \textbf{27.3} \tabularnewline
						
			& \small Pix2PixHD-AUG & \small 54.2 & \small 21.9 & \small 31.8 & \small 19.9 & \small 54.1 & \small 69.2 & \small 47.8 & \small 29.5 & \small \textbf{46.8} & \small 26.0 \tabularnewline
			
			
			\cmidrule(lr){2-12}	
			
			
			\multirow{15}{*}{  \rotatebox{90}{ADE20K }} & \small \small	CC-FPSE   & \small 33.2 & \small 42.6 & \small 44.5 & \small 41.6 & \small 82.2 & \small 87.6 & \small 72.0 & \small 49.8 & \small 52.7 & \small 48.3 \tabularnewline
			
			& \small CC-FPSE-CutMix & \small 33.9 & \small	43.0 & \small	45.1 & \small	41.9 & \small	82.1 & \small	87.2 & \small	72.6 & \small	50.1 & \small	53.4 & \small	48.4 \tabularnewline	
			
			& \small CC-FPSE-Donly & \small \textbf{32.2} & \small	42.9 & \small	45.2 & \small	41.7 & \small	82.5 & \small	87.6 & \small	72.8 & \small	50.3 & \small	54.0 & \small	48.3 \tabularnewline	
			& \small CC-FPSE-Gonly & \small 32.6 & \small	43.7 & \small	44.9 & \small	43.0 & \small	82.5 & \small	87.6 & \small	73.0 & \small	50.8 & \small	53.0 & \small \textbf{49.6} \tabularnewline	
			
			& \small CC-FPSE-AUG & \small 32.6 & \small \textbf{44.0} & \small \textbf{45.8} & \small \textbf{43.1} & \small \textbf{83.0} & \small \textbf{88.2} & \small \textbf{73.5} & \small \textbf{51.3} & \small \textbf{54.5} & \small \textbf{49.6} \tabularnewline[0.2cm]

			
			& \small SPADE & \small 34.4 & \small 39.6 & \small 41.7 & \small 38.6 & \small 81.0 & \small 85.8 & \small 72.1 & \small 46.6 & \small 49.1 & \small 45.3 \tabularnewline
			
			& \small SPADE-CutMix & \small \textbf{34.0} & \small 40.5 & \small 42.8 & \small 39.3 & \small 81.2 & \small \textbf{86.2} & \small 72.0 & \small 47.4 & \small 50.6 & \small 45.7 \tabularnewline
			
			& \small SPADE-Donly & \small 36.0 & \small	39.4 & \small	41.3 & \small	38.4 & \small	81.0 & \small	86.0 & \small	71.6 & \small	46.1 & \small	48.8 & \small	44.7 \tabularnewline
			& \small SPADE-Gonly & \small 35.8 & \small	39.6 & \small 40.9 & \small	38.8 & \small 80.5 & \small	85.6 & \small 71.1 & \small	46.2 & \small 48.3 & \small 45.2 \tabularnewline
			
			& \small SPADE-AUG & \small 34.6 & \small \textbf{41.2} & \small \textbf{43.2} & \small \textbf{40.2} & \small \textbf{81.3} & \small 86.0 & \small \textbf{72.5} & \small \textbf{48.0} & \small \textbf{50.8} & \small \textbf{46.5} \tabularnewline[0.2cm] 
			
			
			& \small Pix2PixHD & \small 59.1 & \small 24.4 & \small 27.9 & \small 22.6 & \small 71.8 & \small 80.5 & \small 55.3 & \small 29.5 & \small 34.7 & \small 26.8 \tabularnewline
			
			& \small Pix2PixHD-CutMix & \small 67.9 & \small 21.7 & \small 25.2 & \small 19.9 & \small 69.8 & \small 78.9 & \small 52.6 & \small 26.3 & \small 31.3 & \small	23.7 \tabularnewline
			
			& \small Pix2PixHD-Donly & \small 42.4 & \small	31.3 & \small 34.7 & \small	29.5 & \small 77.3 & \small	83.8 & \small 65.1 & \small	37.0 & \small 41.4 & \small	34.6 \tabularnewline
			& \small Pix2PixHD-Gonly & \small 63.1 & \small	21.8 & \small 25.5 & \small	19.8 & \small 70.0 & \small	78.9 & \small 53.3 & \small	26.7 & \small 32.2 & \small	23.8 \tabularnewline
			
			& \small Pix2PixHD-AUG & \small \textbf{41.5} & \small \textbf{32.5} & \small \textbf{36.0} & \small \textbf{30.7} & \small \textbf{77.9} & \small \textbf{84.4} & \small \textbf{65.9} & \small \textbf{38.2} & \small \textbf{42.7} & \small \textbf{35.8} \tabularnewline
			
			
			\cmidrule(lr){2-12}  		
			
			\multirow{15}{*}{ \rotatebox{90}{Cityscapes }} & \small CC-FPSE & \small 53.6 & \small 61.8 & \small 79.3 & \small 55.5 & \small 92.8 & \small 94.6 & \small 84.5 & \small 71.6 & \small \textbf{85.6} & \small 66.6 \tabularnewline
			
			& \small CC-FPSE-CutMix & \small 55.2 & \small	62.0 & \small	78.6 & \small	56.1 & \small	92.2 & \small	93.8 & \small	84.8 & \small \textbf{72.7} & \small	85.2 & \small \textbf{68.2} \tabularnewline	
			
			& \small CC-FPSE-Donly & \small 77.7 & \small	53.6 & \small	74.7 & \small	46.0 & \small	89.1 & \small	90.9 & \small	81.3 & \small	64.7 & \small	82.6 & \small	58.4 \tabularnewline
			& \small CC-FPSE-Gonly & \small 57.6 & \small	55.5 & \small	75.9 & \small	48.2 & \small	90.4 & \small	92.1 & \small	82.3 & \small	67.3 & \small	83.2 & \small	61.6 \tabularnewline	
			
			& \small CC-FPSE-AUG & \small \textbf{52.1} & \small \textbf{63.1} & \small \textbf{80.1} & \small \textbf{57.0} & \small \textbf{93.5} & \small \textbf{95.4} & \small \textbf{85.2} & \small 72.4 & \small \textbf{85.6} & \small 67.7 \tabularnewline[0.2cm]

			
			& \small SPADE & \small 64.7 & \small 59.2 & \small \textbf{79.9} & \small 51.9 & \small 93.1 & \small \textbf{95.6} & \small 82.3 & \small 68.3 & \small \textbf{85.5} & \small 62.1 \tabularnewline
			
			& \small SPADE-CutMix & \small \textbf{61.8} & \small 59.9 & \small	79.7 & \small	52.9 & \small \textbf{93.2} & \small 95.5 & \small	82.9 & \small 68.9 & \small	85.2 & \small 63.1 \tabularnewline
			
			& \small SPADE-Donly & \small 63.8 & \small	57.6 & \small	79.8 & \small	49.6 & \small 93.0 & \small	95.5 & \small 81.9 & \small	66.6 & \small \textbf{85.5} & \small 59.8 \tabularnewline
			& \small SPADE-Gonly & \small 64.1 & \small	\textbf{60.5} & \small	79.8 & \small \textbf{53.6} & \small \textbf{93.2} & \small	95.5 & \small \textbf{83.0} & \small 69.3 & \small	85.3 & \small 63.6 \tabularnewline
						
			& \small SPADE-AUG & \small 62.3 & \small 60.4 & \small 79.8 & \small 53.5 & \small \textbf{93.2} & \small 95.5 & \small \textbf{83.0} & \small \textbf{69.7} & \small 85.2 & \small \textbf{64.2} \tabularnewline[0.2cm] 
			
			
			& \small Pix2PixHD & \small 73.6 & \small 56.7& \small 78.8& \small 48.8 & \small \textbf{92.7} & \small \textbf{95.3} & \small 80.7 & \small 65.4 & \small 84.2 & \small 58.7 \tabularnewline
			
			& \small Pix2PixHD-CutMix & \small 83.8 & \small 56.7 & \small	78.7 & \small 48.9 & \small	92.5 & \small	95.2 & \small	80.2 & \small 65.3 & \small	84.2 & \small	58.6 \tabularnewline
			
			& \small Pix2PixHD-Donly & \small 82.4 & \small	56.3 & \small	78.7 & \small 48.3 & \small 92.5 & \small \textbf{95.3} & \small 80.0 & \small	65.1 & \small 84.4 & \small	58.3 \tabularnewline
			& \small Pix2PixHD-Gonly & \small 83.3 & \small	56.5 & \small	78.5 & \small 48.6 & \small	92.5 & \small 95.2 & \small	80.4 & \small	65.3 & \small 84.0 & \small 58.6 \tabularnewline	
			
			& \small Pix2PixHD-AUG & \small \textbf{72.7} & \small \textbf{58.0} & \small \textbf{79.1} & \small \textbf{50.5} & \small \textbf{92.7} & \small \textbf{95.3} & \small \textbf{81.2} & \small \textbf{66.9} & \small \textbf{84.6} & \small \textbf{60.5} \tabularnewline 
		\end{tabular}%
	}
	\vspace{0.5em}
	\caption{Ablation results of each model on COCO-Stuff, ADE20K and Cityscapes. ${BC}$ and ${UC}$ denote the biased and unbiased class splits. \textbf{Bold} indicates the best model. For Donly and Gonly models, the augmentation was applied only to the discriminator and the generator input, respectively. For the ease of comparison, we repeat the metrics from the respective baseline and augmented models (-AUG)}
	\label{tab:cutmix_selective_augmentation_ablations}
\end{table*}
\clearpage

%% file: tables/ablation_edge_loss.tex
\begin{table*}[h]
	\setlength{\tabcolsep}{0.2em}
	\renewcommand{\arraystretch}{0.95}

	\centering
	\resizebox{0.9\linewidth}{!}{%
		\begin{tabular}{@{}lc@{\hspace{0.2cm}}c@{\hspace{0.5cm}}ccc@{\hspace{0.5cm}}ccc@{\hspace{0.5cm}}ccc@{}}
			
			& \small  Method& \small FID $\downarrow$& \small mIoU $\uparrow$& \small mIoU$_{BC}$& \small mIoU$_{UC}$& \small PA $\uparrow$& \small PA$_{BC}$& \small PA$_{UC}$& \small MA $\uparrow$& \small MA$_{BC}$& \small MA$_{UC}$ \tabularnewline  
			\cmidrule(lr){2-12} 
			
			& \small CC-FPSE-AUG-woEL & 53.4 & 63.1 & \textbf{80.4} & 56.9 & \textbf{93.6} & 95.4 & \textbf{85.5} & 72.4 & \textbf{86.0} & 67.6 \tabularnewline 
			& \small CC-FPSE-AUG &  \textbf{52.1} &  63.1 &  80.1 &  \textbf{57.0} &  93.5 &  95.4 & 85.2 & 72.4 & 85.6 &  \textbf{67.7} \tabularnewline [0.2cm]
			
			&  \small SPADE-AUG-woEL &  \textbf{33.8} &  40.3 &  42.2 &  39.3 &  80.9 &  85.7 &  71.9 &  47.3 &  49.8 &  45.9 \tabularnewline 
			& \small SPADE-AUG &  34.6 &  \textbf{41.2} &  \textbf{43.2} &  \textbf{40.2} &  \textbf{81.3} &  \textbf{86.0} &  \textbf{72.5} &  \textbf{48.0} &  \textbf{50.8} &  \textbf{46.5} \tabularnewline [0.2cm] 
			
			&  \small Pix2PixHD-AUG-woEL &  137.2 &  14.5 &  23.1 &  12.7 &  39.9 &  52.7 &  34.6 &  21.4 &  33.2 &  19.0 \tabularnewline 
			& \small 	Pix2PixHD-AUG& \small \textbf{54.2} & \small \textbf{21.9} & \small \textbf{31.8} & \small \textbf{19.9} & \small \textbf{54.1} & \small \textbf{69.2} & \small \textbf{47.8} & \small \textbf{29.5} & \small \textbf{46.8} & \small \textbf{26.0} \tabularnewline
			
		\end{tabular}%
		}
	\vspace{0.5em}
	\caption{Augmented models without the edge loss (AUG-woEL). The CC-FPSE, SPADE and Pix2PixHD models are trained on Cityscapes, ADE20k and COCO-Stuff, respectively. ${BC}$ and ${UC}$ denote the biased and unbiased class splits. For the ease of comparison, we repeat the metrics from the respective augmented models (-AUG). The augmented models refer to the baseline models trained with the full augmentation scheme proposed in this work (warping + edge loss).
	}
	\label{tab:edge_loss_ablation}
\end{table*}

%% file: tables/matching_aware_loss.tex
\begin{table*}[h]
	\setlength{\tabcolsep}{0.2em}
	\renewcommand{\arraystretch}{0.95}
	\centering

	\centering
	\resizebox{0.9\linewidth}{!}{%
		\begin{tabular}{@{}lc@{\hspace{0.2cm}}c@{\hspace{0.5cm}}ccc@{\hspace{0.5cm}}ccc@{\hspace{0.5cm}}ccc@{}}
			
			& \small  Method& \small FID $\downarrow$& \small mIoU $\uparrow$& \small mIoU$_{BC}$& \small mIoU$_{UC}$& \small PA $\uparrow$& \small PA$_{BC}$& \small PA$_{UC}$& \small MA $\uparrow$& \small MA$_{BC}$& \small MA$_{UC}$ \tabularnewline 
			\cmidrule(lr){2-12} 
			& \small CC-FPSE-AUG-MAL & 35.4 & 42.7 & 44.8 & 41.6 & 82.3 & 87.4 & 72.6 & 50.2 & 53.4 & 48.5 \tabularnewline 
			& \small CC-FPSE-AUG & \textbf{32.6} & \textbf{44.0} & \textbf{45.8} & \textbf{43.1} & \textbf{83.0} & \textbf{88.2} & \textbf{73.5} & \textbf{51.3} & \textbf{54.5} & \textbf{49.6} \tabularnewline[0.2cm]
			 	
			& \small SPADE-AUG-MAL & 35.6 & 39.5 & 41.4 & 38.5 & 80.9 & 85.7 & 71.9 & 46.2 & 48.7 & 44.9 \tabularnewline 
			& \small SPADE-AUG & \textbf{34.6} & \textbf{41.2} & \textbf{43.2} & \textbf{40.2} & \textbf{81.3} & \textbf{86.0} & \textbf{72.5} & \textbf{48.0} & \textbf{50.8} & \textbf{46.5} \tabularnewline[0.2cm] 
			
			& \small Pix2PixHD-AUG-MAL & 69.4 & 23.7 & 26.7 & 22.1 & 72.1 &	80.8 & 56.0 & 28.8 & 33.5 & 26.2 \tabularnewline
			& \small Pix2PixHD-AUG & \textbf{40.7} & \textbf{33.1} & \textbf{36.3} & \textbf{31.5} & \textbf{77.5} & \textbf{84.0} & \textbf{65.4} & \textbf{39.2} & \textbf{43.2} & \textbf{37.0} \tabularnewline
			
		\end{tabular}%
	}
	\vspace{0.5em}
	\caption{Scores obtained using augmented models with the matching aware GAN loss (AUG-MAL) on ADE20K dataset. ${BC}$ and ${UC}$ denote the biased and unbiased class splits. For all baselines, this extension did not improve the performance of the augmented models. For the ease of comparison, we repeat the metrics from the respective augmented models (-AUG). The augmented models refer to the baseline models trained with the full augmentation scheme proposed in this work (warping + edge loss).}
	\label{tab:matching_aware_loss_ablation}
\end{table*}

%% file: figures/ccfpse_coco.tex
\begin{figure*}[h]
	\centering
	\begin{minipage}[c]{1.0\linewidth}
		\centering
		\setlength{\tabcolsep}{0em}
		\renewcommand{\arraystretch}{0}

		\hfill{}%
\begin{tabular}{@{}c@{}c@{}c@{}c@{}c@{}}
			
			\centering
			
			Label map & Real image & CC-FPSE & CC-FPSE-EL & CC-FPSE-AUG \vspace{0.05cm} \tabularnewline
			
			\includegraphics[width=0.19\linewidth, height=0.09\textheight]{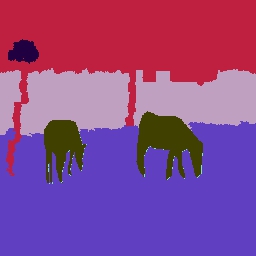} & {\footnotesize{}}
			\includegraphics[width=0.19\linewidth, height=0.09\textheight]{figures/models/ccfpse/coco/org_img/000000020059.jpg} & {\footnotesize{}}
			\includegraphics[width=0.19\linewidth, height=0.09\textheight]{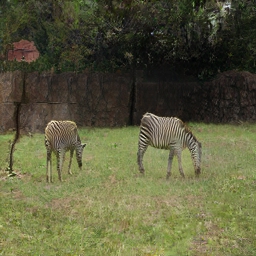} & {\footnotesize{}}
			\includegraphics[width=0.19\linewidth, height=0.09\textheight]{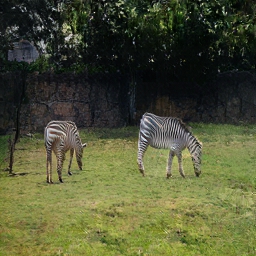} & {\footnotesize{}}
			\includegraphics[width=0.19\linewidth, height=0.09\textheight]{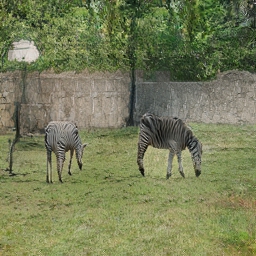} \vspace{0.05cm} \tabularnewline
			
			\includegraphics[width=0.19\linewidth, height=0.09\textheight]{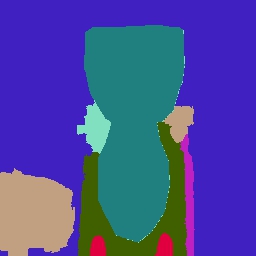} & {\footnotesize{}}
			\includegraphics[width=0.19\linewidth, height=0.09\textheight]{figures/models/ccfpse/coco/org_img/000000038070.jpg} & {\footnotesize{}}
			\includegraphics[width=0.19\linewidth, height=0.09\textheight]{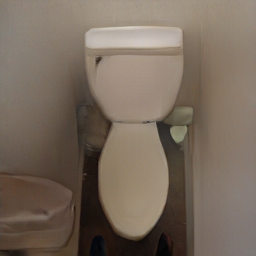} & {\footnotesize{}}
			\includegraphics[width=0.19\linewidth, height=0.09\textheight]{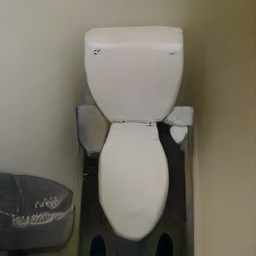} & {\footnotesize{}}
			\includegraphics[width=0.19\linewidth, height=0.09\textheight]{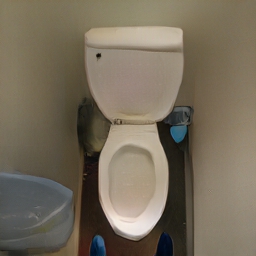} \vspace{0.05cm} \tabularnewline
\end{tabular}
	\end{minipage}
	\vspace{0.2em}
	\caption{COCO-Stuff results obtained using CC-FPSE baselines and the augmented model. Best viewed in color. The red boxes highlight image artifacts.}
	\label{fig:ccfpse_coco}
\end{figure*}

%% file: figures/ccfpse_ade20k.tex
\begin{figure*}[h]
	\centering
	\begin{minipage}[c]{1.0\linewidth}
		\centering
		\setlength{\tabcolsep}{0em}
		\renewcommand{\arraystretch}{0}

		\hfill{}%
		\begin{tabular}{@{}c@{}c@{}c@{}c@{}c@{}}
			
			\centering
			
			Label map & Real image & CC-FPSE & CC-FPSE-EL & CC-FPSE-AUG \vspace{0.05cm} \tabularnewline
			
			\includegraphics[width=0.19\linewidth, height=0.09\textheight]{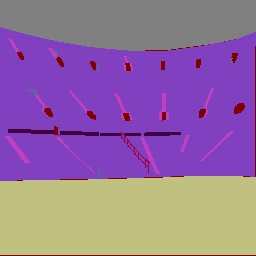} & {\footnotesize{}}
			\includegraphics[width=0.19\linewidth, height=0.09\textheight]{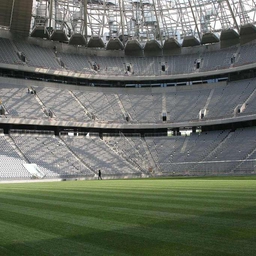} & {\footnotesize{}}
			\includegraphics[width=0.19\linewidth, height=0.09\textheight]{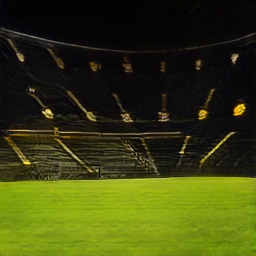} & {\footnotesize{}}
			\includegraphics[width=0.19\linewidth, height=0.09\textheight]{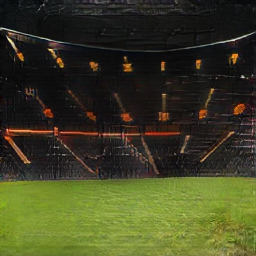} & {\footnotesize{}}
			\includegraphics[width=0.19\linewidth, height=0.09\textheight]{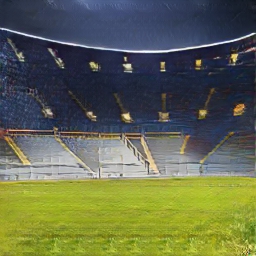} \vspace{0.05cm} \tabularnewline
			
			\includegraphics[width=0.19\linewidth, height=0.09\textheight]{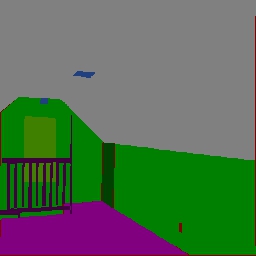} & {\footnotesize{}}
			\includegraphics[width=0.19\linewidth, height=0.09\textheight]{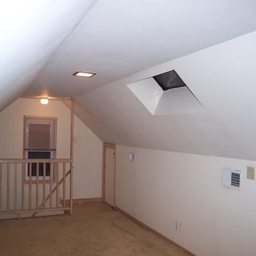} & {\footnotesize{}}
			\includegraphics[width=0.19\linewidth, height=0.09\textheight]{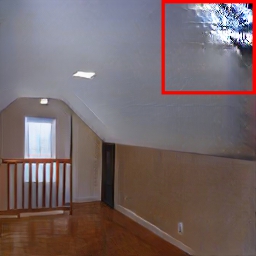} & {\footnotesize{}}
			\includegraphics[width=0.19\linewidth, height=0.09\textheight]{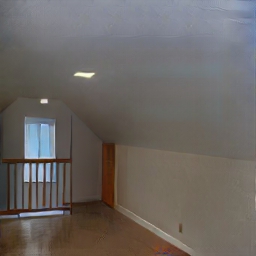} & {\footnotesize{}}
			\includegraphics[width=0.19\linewidth, height=0.09\textheight]{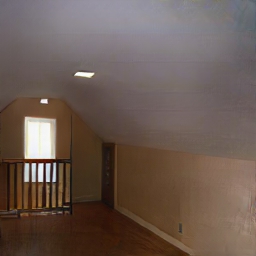} \vspace{0.05cm} \tabularnewline
\end{tabular}
	\end{minipage}
	\vspace{0.2em}
	\caption{ADE20K results obtained using CC-FPSE baselines and the augmented model. Best viewed in color. The red boxes highlight image artifacts.}
	\label{fig:ccfpse_ade20k}
\end{figure*}

%% file: figures/ccfpse_cityscapes.tex
\begin{figure*}[h]
	\centering
	\setlength{\tabcolsep}{0em}
	\renewcommand{\arraystretch}{0}
	\hfill{}%
	\begin{tabular}{@{}c@{}c@{}c@{}}
		
		\centering
		
		CC-FPSE & CC-FPSE-EL & CC-FPSE-AUG \vspace{0.05cm} \tabularnewline
		
		\includegraphics[width=0.32\linewidth]{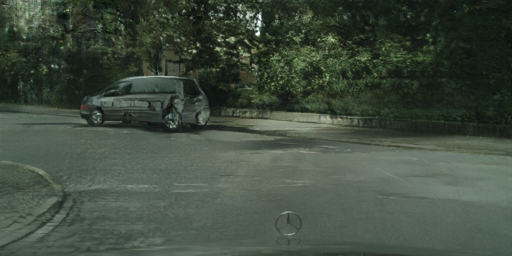} & {\footnotesize{}}
		\includegraphics[width=0.32\linewidth]{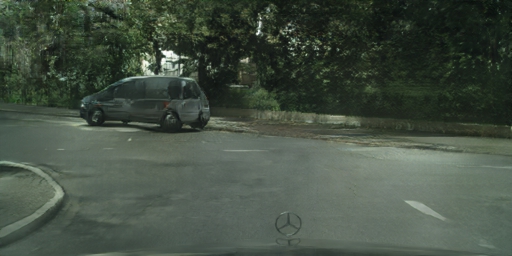} & {\footnotesize{}}
		\includegraphics[width=0.32\linewidth]{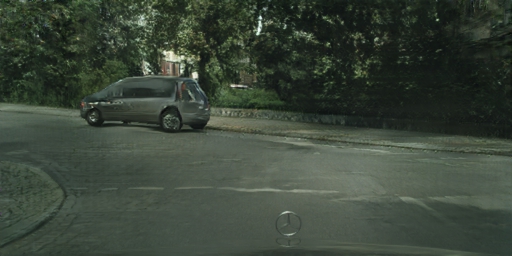} \vspace{0.05cm} \tabularnewline
		
		\includegraphics[width=0.32\linewidth]{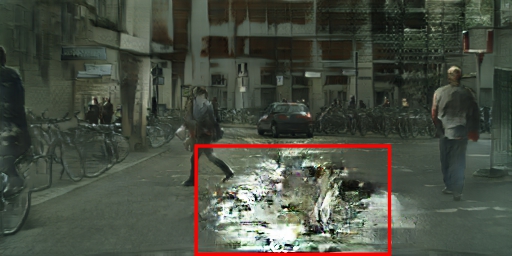} & {\footnotesize{}}
		\includegraphics[width=0.32\linewidth]{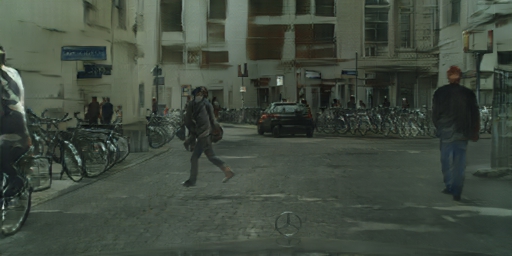} & {\footnotesize{}}
		\includegraphics[width=0.32\linewidth]{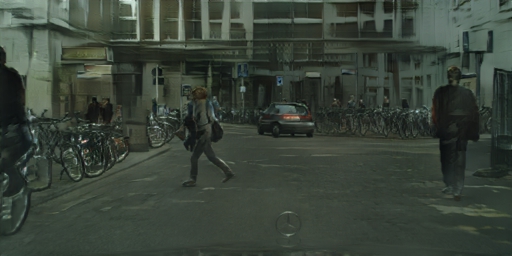} \vspace{0.05cm} \tabularnewline 
		
		\includegraphics[width=0.32\linewidth]{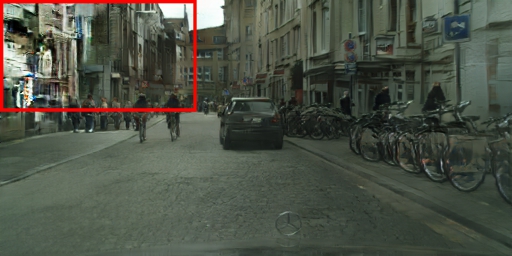} & {\footnotesize{}}
		\includegraphics[width=0.32\linewidth]{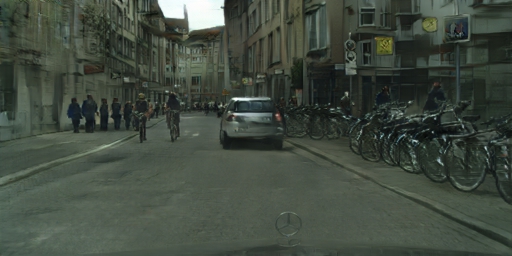} & {\footnotesize{}}
		\includegraphics[width=0.32\linewidth]{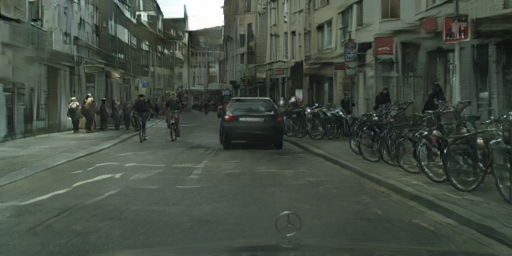} \vspace{0.05cm} \tabularnewline
		
		\includegraphics[width=0.32\linewidth]{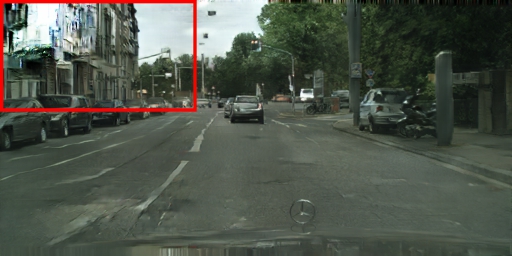} & {\footnotesize{}}
		\includegraphics[width=0.32\linewidth]{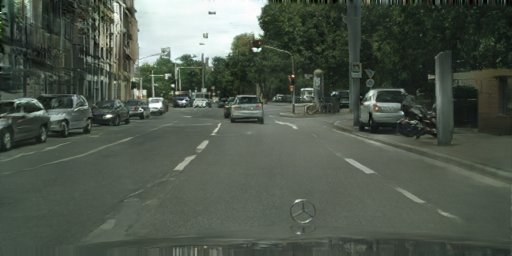} & {\footnotesize{}}
		\includegraphics[width=0.32\linewidth]{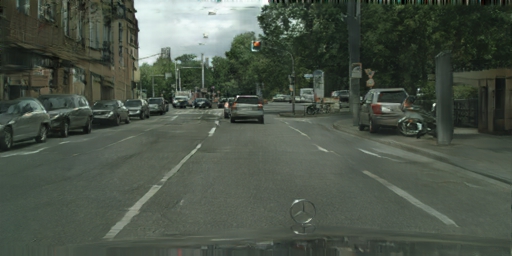} \vspace{0.05cm} \tabularnewline
\end{tabular}
	\vspace{0.2em}
	\caption{Cityscapes results obtained using CC-FPSE baselines and the augmented model. For clarity, we omit the label maps and the real images. Best viewed in color. The red boxes highlight image artifacts.}
	\label{fig:ccfpse_cityscapes}
\end{figure*}

%% file: figures/spade_coco.tex
\begin{figure*}[t]
	\centering
	\begin{minipage}[c]{1.0\linewidth}
		\centering
		\setlength{\tabcolsep}{0em}
		\renewcommand{\arraystretch}{0}

		\hfill{}%
		\begin{tabular}{@{}c@{}c@{}c@{}c@{}c@{}}
			
			\centering
			
			Label map & Real image & SPADE & SPADE-EL & SPADE-AUG \vspace{0.05cm} \tabularnewline
			
			\includegraphics[width=0.19\linewidth, height=0.09\textheight]{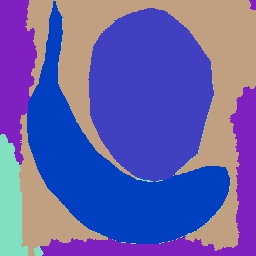} & {\footnotesize{}}
			\includegraphics[width=0.19\linewidth, height=0.09\textheight]{figures/models/spade/coco/org_img/000000002587.jpg} & {\footnotesize{}}
			\includegraphics[width=0.19\linewidth, height=0.09\textheight]{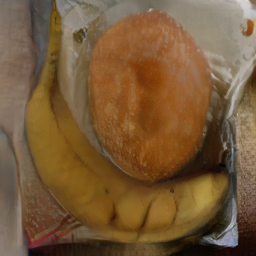} & {\footnotesize{}}
			\includegraphics[width=0.19\linewidth, height=0.09\textheight]{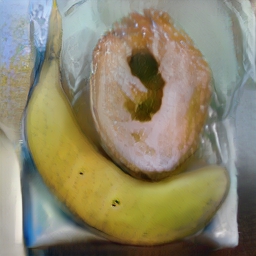} & {\footnotesize{}}
			\includegraphics[width=0.19\linewidth, height=0.09\textheight]{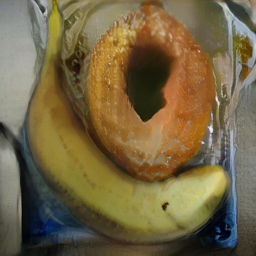} \vspace{0.05cm} \tabularnewline
			
			\includegraphics[width=0.19\linewidth, height=0.09\textheight]{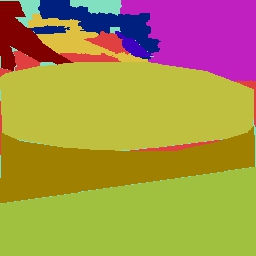} & {\footnotesize{}}
			\includegraphics[width=0.19\linewidth, height=0.09\textheight]{figures/models/spade/coco/org_img/000000032038.jpg} & {\footnotesize{}}
			\includegraphics[width=0.19\linewidth, height=0.09\textheight]{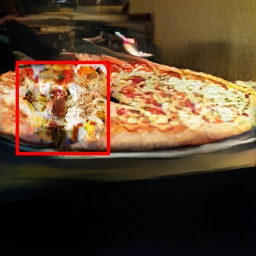} & {\footnotesize{}}
			\includegraphics[width=0.19\linewidth, height=0.09\textheight]{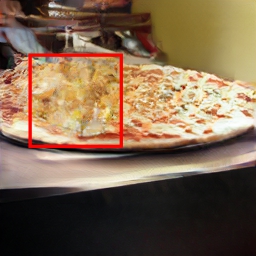} & {\footnotesize{}}
			\includegraphics[width=0.19\linewidth, height=0.09\textheight]{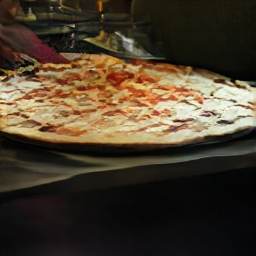} \vspace{0.05cm} \tabularnewline
\end{tabular}
	\end{minipage}
	\vspace{0.2em}
	\caption{COCO-Stuff results obtained using SPADE baselines and the augmented model. Best viewed in color. The red boxes highlight image artifacts.}
	\label{fig:spade_coco}
\end{figure*}

%% file: figures/spade_ade20k.tex
\begin{figure*}[t]
	\centering
	\begin{minipage}[c]{1.0\linewidth}
		\centering
		\setlength{\tabcolsep}{0em}
		\renewcommand{\arraystretch}{0}

		\hfill{}%
		\begin{tabular}{@{}c@{}c@{}c@{}c@{}c@{}}
			
			\centering
			
			Label map & Real image & SPADE & SPADE-EL & SPADE-AUG \vspace{0.05cm} \tabularnewline
			
			\includegraphics[width=0.19\linewidth, height=0.09\textheight]{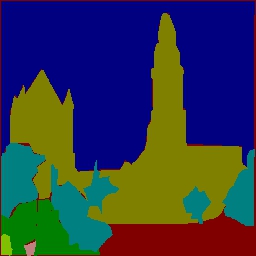} & {\footnotesize{}}
			\includegraphics[width=0.19\linewidth, height=0.09\textheight]{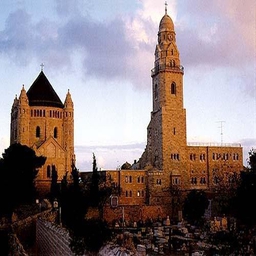} & {\footnotesize{}}
			\includegraphics[width=0.19\linewidth, height=0.09\textheight]{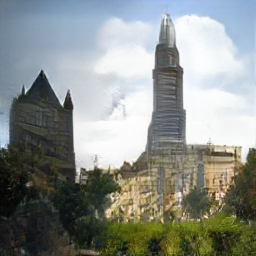} & {\footnotesize{}}
			\includegraphics[width=0.19\linewidth, height=0.09\textheight]{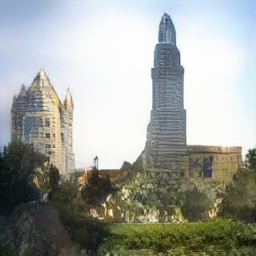} & {\footnotesize{}}
			\includegraphics[width=0.19\linewidth, height=0.09\textheight]{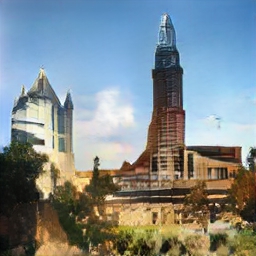} \vspace{0.05cm} \tabularnewline
			
			\includegraphics[width=0.19\linewidth, height=0.09\textheight]{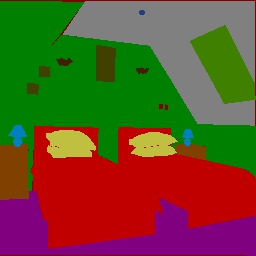} & {\footnotesize{}}
			\includegraphics[width=0.19\linewidth, height=0.09\textheight]{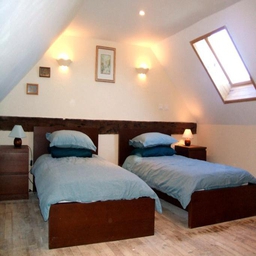} & {\footnotesize{}}
			\includegraphics[width=0.19\linewidth, height=0.09\textheight]{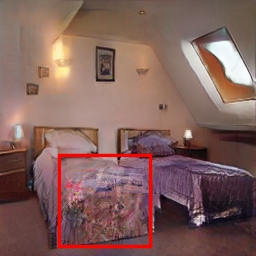} & {\footnotesize{}}
			\includegraphics[width=0.19\linewidth, height=0.09\textheight]{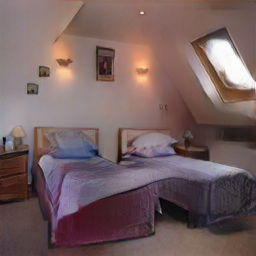} & {\footnotesize{}}
			\includegraphics[width=0.19\linewidth, height=0.09\textheight]{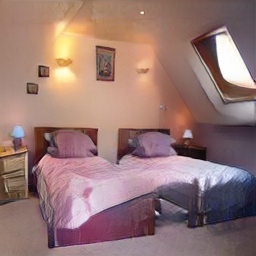} \vspace{0.05cm} \tabularnewline
\end{tabular}
	\end{minipage}
	\vspace{0.2em}
	\caption{ADE20K results obtained using SPADE baselines and the augmented model. Best viewed in color. The red boxes highlight image artifacts.}
	\label{fig:spade_ade20k}
\end{figure*}

%% file: figures/spade_cityscapes.tex
\begin{figure*}[t]
	\centering
	\setlength{\tabcolsep}{0em}
	\renewcommand{\arraystretch}{0}
	\hfill{}%
	\begin{tabular}{@{}c@{}c@{}c@{}}
		
		\centering
		
		SPADE & SPADE-EL & SPADE-AUG \vspace{0.05cm} \tabularnewline
		
		\includegraphics[width=0.32\linewidth]{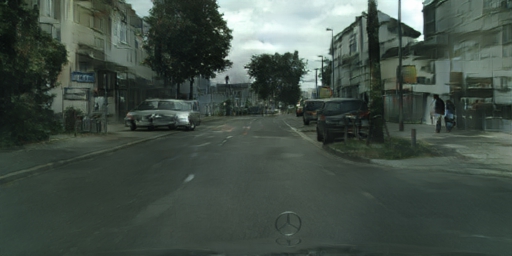} & {\footnotesize{}}
		\includegraphics[width=0.32\linewidth]{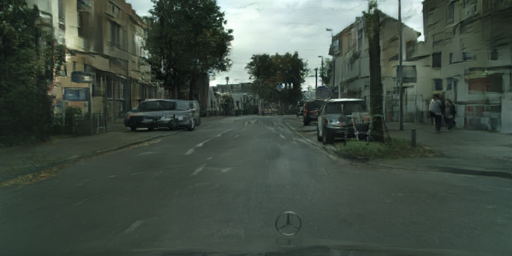} & {\footnotesize{}}
		\includegraphics[width=0.32\linewidth]{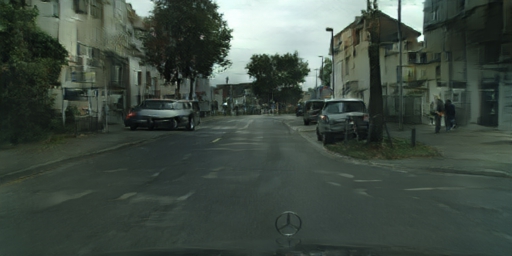} \vspace{0.05cm} \tabularnewline
		
		\includegraphics[width=0.32\linewidth]{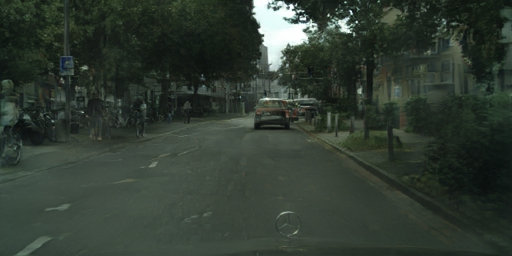} & {\footnotesize{}}
		\includegraphics[width=0.32\linewidth]{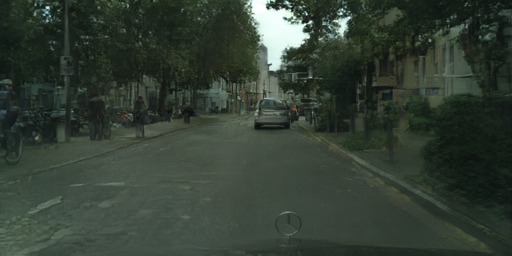} & {\footnotesize{}}
		\includegraphics[width=0.32\linewidth]{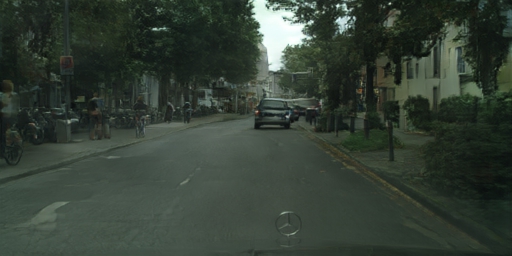} \vspace{0.05cm} \tabularnewline 
		
		\includegraphics[width=0.32\linewidth]{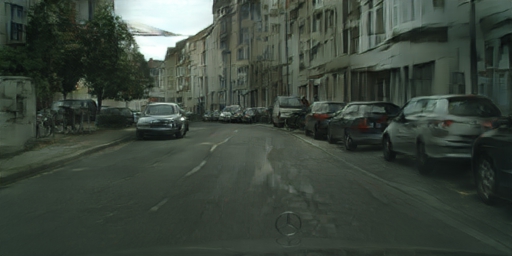} & {\footnotesize{}}
		\includegraphics[width=0.32\linewidth]{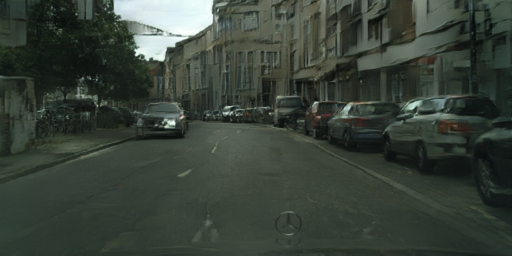} & {\footnotesize{}}
		\includegraphics[width=0.32\linewidth]{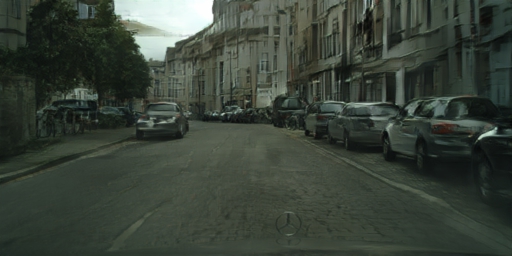} \vspace{0.05cm} \tabularnewline
		
		\includegraphics[width=0.32\linewidth]{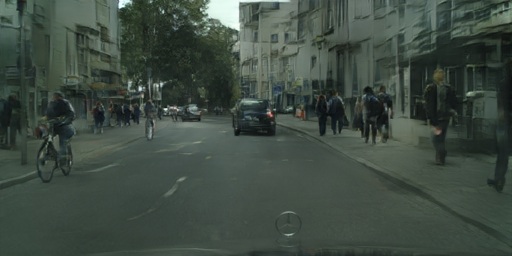} & {\footnotesize{}}
		\includegraphics[width=0.32\linewidth]{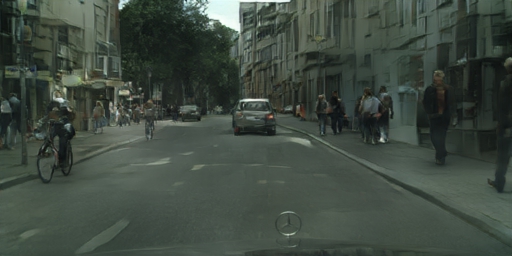} & {\footnotesize{}}
		\includegraphics[width=0.32\linewidth]{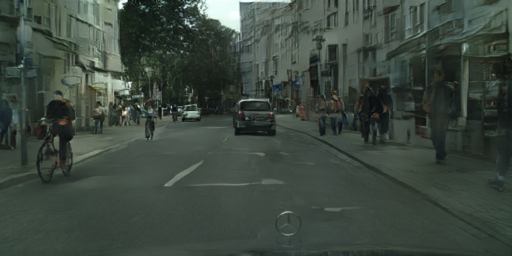} \vspace{0.05cm} \tabularnewline
\end{tabular}
	\vspace{0.2em}
	\caption{Cityscapes results obtained using SPADE baselines and the augmented model. For clarity, we omit the label maps and the real images. Best viewed in color.}
	\label{fig:spade_cityscapes}
\end{figure*}

%% file: figures/pix2pixhd_coco.tex
\begin{figure*}[t]
	\centering
	\begin{minipage}[c]{1.0\linewidth}
		\centering
		\setlength{\tabcolsep}{0em}
		\renewcommand{\arraystretch}{0}

		\hfill{}%
		\begin{tabular}{@{}c@{}c@{}c@{}c@{}c@{}}
			
			\centering
			
			Label map & Real image & Pix2PixHD & Pix2PixHD-EL & Pix2PixHD-AUG \vspace{0.05cm} \tabularnewline
			
			\includegraphics[width=0.19\linewidth, height=0.09\textheight]{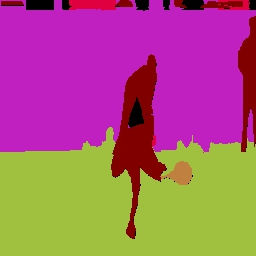} & {\footnotesize{}}
			\includegraphics[width=0.19\linewidth, height=0.09\textheight]{figures/models/pix2pixhd/coco/org_img/000000000885.jpg} & {\footnotesize{}}
			\includegraphics[width=0.19\linewidth, height=0.09\textheight]{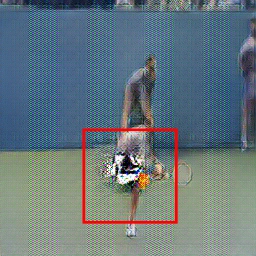} & {\footnotesize{}}
			\includegraphics[width=0.19\linewidth, height=0.09\textheight]{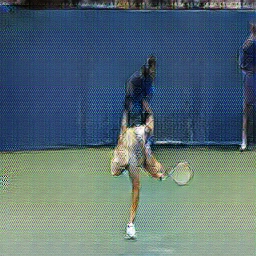} & {\footnotesize{}}
			\includegraphics[width=0.19\linewidth, height=0.09\textheight]{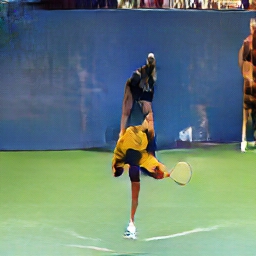} \vspace{0.05cm} \tabularnewline
			
			\includegraphics[width=0.19\linewidth, height=0.09\textheight]{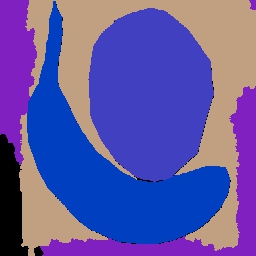} & {\footnotesize{}}
			\includegraphics[width=0.19\linewidth, height=0.09\textheight]{figures/models/pix2pixhd/coco/org_img/000000002587.jpg} & {\footnotesize{}}
			\includegraphics[width=0.19\linewidth, height=0.09\textheight]{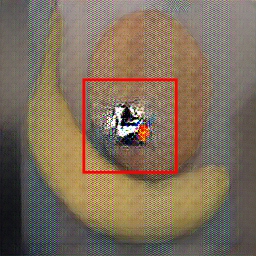} & {\footnotesize{}}
			\includegraphics[width=0.19\linewidth, height=0.09\textheight]{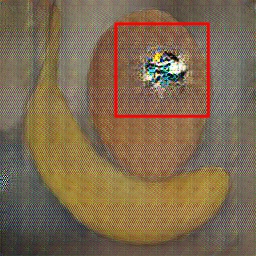} & {\footnotesize{}}
			\includegraphics[width=0.19\linewidth, height=0.09\textheight]{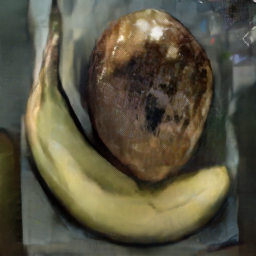} \vspace{0.05cm} \tabularnewline
\end{tabular}
	\end{minipage}
	\vspace{0.2em}
	\caption{COCO-Stuff results obtained using Pix2PixHD baselines and the augmented model. Best viewed in color. The red boxes highlight image artifacts.}
	\label{fig:pix2pixhd_coco}
\end{figure*}

%% file: figures/pix2pixhd_ade20k.tex
\begin{figure*}[t]
	\centering
	\begin{minipage}[c]{1.0\linewidth}
		\centering
		\setlength{\tabcolsep}{0em}
		\renewcommand{\arraystretch}{0}

		\hfill{}%
		\begin{tabular}{@{}c@{}c@{}c@{}c@{}c@{}}
			
			\centering
			
			Label map & Real image & Pix2PixHD & Pix2PixHD-EL & Pix2PixHD-AUG \vspace{0.05cm} \tabularnewline
			
			\includegraphics[width=0.19\linewidth, height=0.09\textheight]{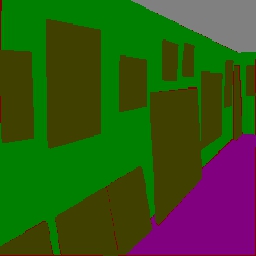} & {\footnotesize{}}
			\includegraphics[width=0.19\linewidth, height=0.09\textheight]{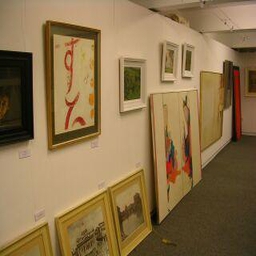} & {\footnotesize{}}
			\includegraphics[width=0.19\linewidth, height=0.09\textheight]{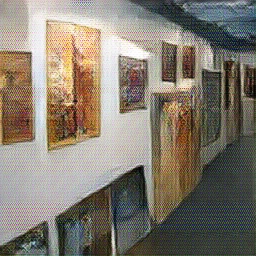} & {\footnotesize{}}
			\includegraphics[width=0.19\linewidth, height=0.09\textheight]{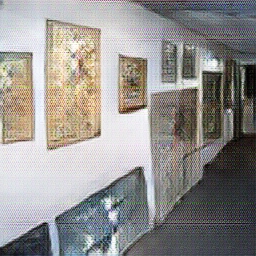} & {\footnotesize{}}
			\includegraphics[width=0.19\linewidth, height=0.09\textheight]{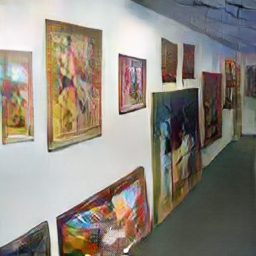} \vspace{0.05cm} \tabularnewline
			
			\includegraphics[width=0.19\linewidth, height=0.09\textheight]{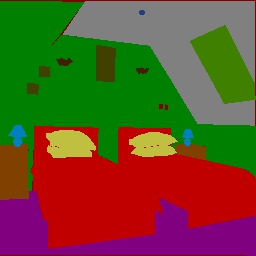} & {\footnotesize{}}
			\includegraphics[width=0.19\linewidth, height=0.09\textheight]{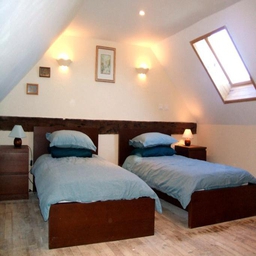} & {\footnotesize{}}
			\includegraphics[width=0.19\linewidth, height=0.09\textheight]{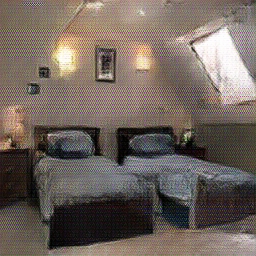} & {\footnotesize{}}
			\includegraphics[width=0.19\linewidth, height=0.09\textheight]{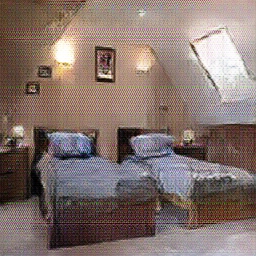} & {\footnotesize{}}
			\includegraphics[width=0.19\linewidth, height=0.09\textheight]{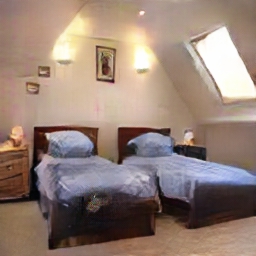} \vspace{0.05cm} \tabularnewline
\end{tabular}
	\end{minipage}
	\vspace{0.2em}
	\caption{ADE20K results obtained using Pix2PixHD baselines and the augmented model. Best viewed in color.}
	\label{fig:pix2pixhd_ade20k}
\end{figure*}

%% file: figures/pix2pixhd_cityscapes.tex
\begin{figure*}[t]
	\centering
	\setlength{\tabcolsep}{0em}
	\renewcommand{\arraystretch}{0}
	\hfill{}%
	\begin{tabular}{@{}c@{}c@{}c@{}}
		
		\centering
		
		Pix2PixHD & Pix2PixHD-EL & Pix2PixHD-AUG \vspace{0.05cm} \tabularnewline
		
		\includegraphics[width=0.32\linewidth]{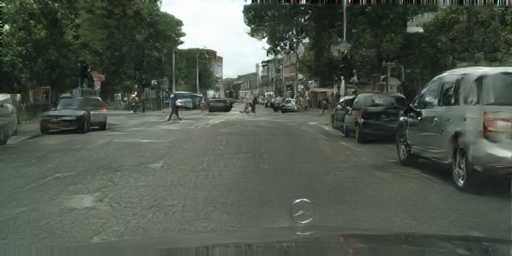} & {\footnotesize{}}
		\includegraphics[width=0.32\linewidth]{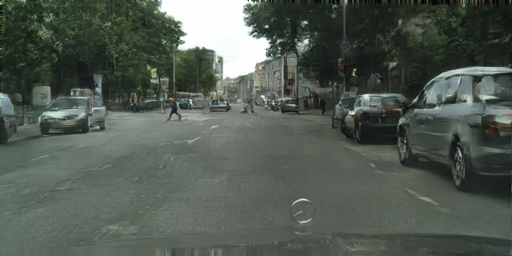} & {\footnotesize{}}
		\includegraphics[width=0.32\linewidth]{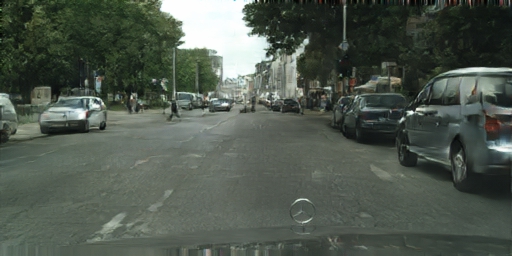} \vspace{0.05cm} \tabularnewline
		
		\includegraphics[width=0.32\linewidth]{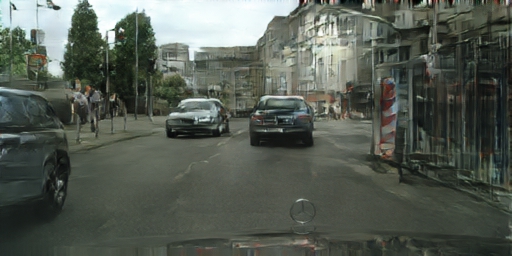} & {\footnotesize{}}
		\includegraphics[width=0.32\linewidth]{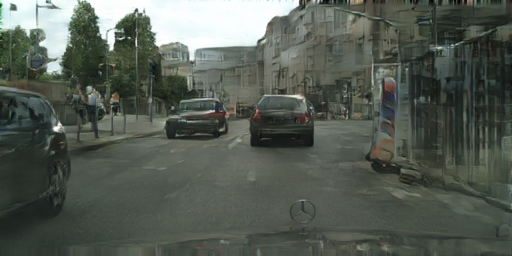} & {\footnotesize{}}
		\includegraphics[width=0.32\linewidth]{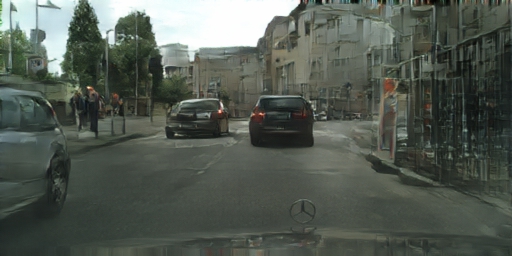} \vspace{0.05cm} \tabularnewline 
		
		\includegraphics[width=0.32\linewidth]{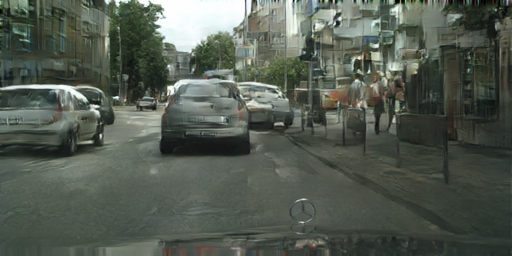} & {\footnotesize{}}
		\includegraphics[width=0.32\linewidth]{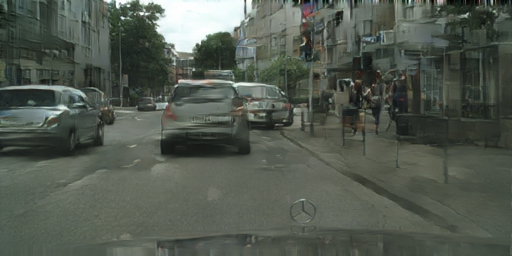} & {\footnotesize{}}
		\includegraphics[width=0.32\linewidth]{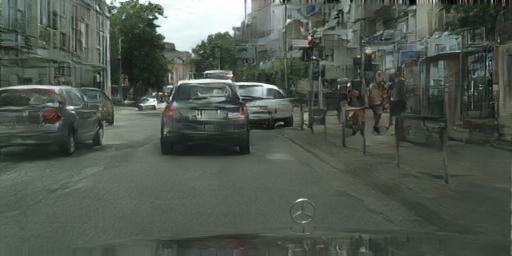} \vspace{0.05cm} \tabularnewline
		
		\includegraphics[width=0.32\linewidth]{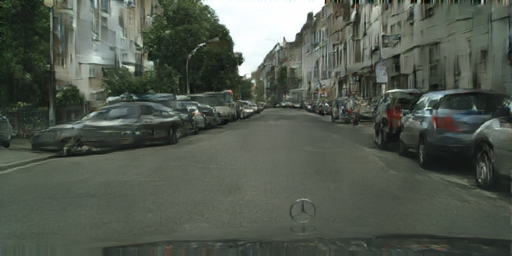} & {\footnotesize{}}
		\includegraphics[width=0.32\linewidth]{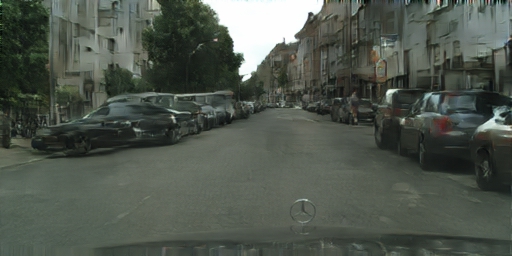} & {\footnotesize{}}
		\includegraphics[width=0.32\linewidth]{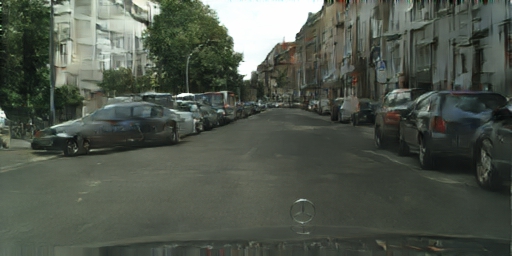} \vspace{0.05cm} \tabularnewline
\end{tabular}
	\vspace{0.2em}
	\caption{Cityscapes results obtained using Pix2PixHD baselines and the augmented model. For clarity, we omit the label maps and the real images. Best viewed in color.}
	\label{fig:pix2pixhd_cityscapes}
\end{figure*}